\documentclass[10pt, a4paper]{article}

\usepackage[]{lrec-coling2024} % this is the new style

\usepackage{silence}
\usepackage{fontawesome5}
\newcommand{\DoCS}{\faLaptopCode}
\newcommand{\IACS}{\faUniversity}

\usepackage{tikz}
\newlength{\uppercaseHeight}
\newlength{\uppercaseWidth}
\newcommand{\DoL}{%
\settoheight{\uppercaseHeight}{\faHeadSideCough}%
\settowidth{\uppercaseWidth}{\faHeadSideCough}%
\faHeadSideCough% Print R
\tikz[baseline,overlay] \fill [white] % [red] % useful to use a red box when tweaking
(-0.23\uppercaseWidth, -0.15*\uppercaseHeight) rectangle ++(0.24*\uppercaseWidth, 0.6*\uppercaseHeight);
\tikz[baseline,overlay] \fill [white] % [red] % useful to use a red box when tweaking
(-0.53\uppercaseWidth, 0.05*\uppercaseHeight) rectangle ++(0.2*\uppercaseWidth, 0.2*\uppercaseHeight);
\hspace{-1.1em}
}

\usepackage{booktabs,multirow}
\usepackage{colortbl}
\newcolumntype{C}[1]{>{\centering\let\newline\\\arraybackslash\hspace{0pt}}m{#1}}

\usepackage{arydshln}
\makeatletter
\def\adl@drawiv#1#2#3{%
        \hskip.5\tabcolsep
        \xleaders#3{#2.5\@tempdimb #1{1}#2.5\@tempdimb}%
                #2\z@ plus1fil minus1fil\relax
        \hskip.5\tabcolsep}
\newcommand{\cdashlinelr}[1]{%
  \noalign{\vskip\aboverulesep
           \global\let\@dashdrawstore\adl@draw
           \global\let\adl@draw\adl@drawiv}
  \cdashline{#1}
  \noalign{\global\let\adl@draw\@dashdrawstore
           \vskip\belowrulesep}}
\makeatother

\usepackage{graphicx}
\usepackage{subcaption}
\usepackage{setspace}

\newcommand{\tsc}[1]{\textsc{#1}}
\newcommand{\ttt}[1]{\texttt{#1}}
\newcommand{\tbf}[1]{\textbf{#1}}
\newcommand{\tss}[1]{\textsuperscript{#1}}

\usepackage{xspace}
\newcommand{\flabel}[1]{\textsc{#1}\xspace}
\newcommand{\SPosR}{\flabel{SPos+}}
\newcommand{\SPosT}{\flabel{SPos-}}
\newcommand{\HPosR}{\flabel{HPos+}}
\newcommand{\HPosT}{\flabel{HPos-}}
\newcommand{\SNegR}{\flabel{SNeg+}}
\newcommand{\SNegT}{\flabel{SNeg-}}
\newcommand{\HNegR}{\flabel{HNeg+}}
\newcommand{\HNegT}{\flabel{HNeg-}}
\newcommand{\Other}{\flabel{Other}}

\newcommand{\FOS}{\flabel{Fos}}
\newcommand{\TAM}{\flabel{Ta-Mrda}}
\newcommand{\TAS}{\flabel{Ta-Swda}}
\newcommand{\MTS}{\flabel{Mtl-Swda}}
\newcommand{\MTM}{\flabel{Mtl-Mrda}}

\newcommand{\sdag}{\tss{\textdagger}}
\newcommand{\sddag}{\tss{\textdaggerdbl}}

\usepackage{enumitem}
\setlist[enumerate]{
    topsep=0pt,itemsep=-0.5ex,
    partopsep=0.5ex,parsep=0.5ex
}
\setlist[itemize]{
    topsep=0pt,itemsep=-0.5ex,
    partopsep=0.5ex,parsep=0.5ex,
}

\newcommand{\email}[1]{\ttt{#1}}

\usepackage{todonotes}

\definecolor{ocr}{HTML}{009900}

\definecolor{as}{HTML}{0000FF}

\title{Intention and Face in Dialog}

\name{
    \tss{\DoCS\IACS}Adil Soubki,
    \tss{\DoL\IACS}Owen Rambow
}

\address{
    \tss{\DoCS}Department of Computer Science,
    \tss{\DoL}Department of Linguistics \\
    \tss{\IACS}Institute for Advanced Computational Science,
    Stony Brook University \\
    \email{asoubki@cs.stonybrook.edu}, \email{owen.rambow@stonybrook.edu}
}

\abstract{
The notion of face described by Brown and Levinson (1987) has been studied in great detail, but a critical aspect of the framework, that which focuses on how intentions mediate the planning of turns which impose upon face, has received far less attention. We present an analysis of three computational systems trained for classifying both intention and politeness, focusing on how the former influences the latter. In politeness theory, agents attend to the desire to have their wants appreciated (positive face), and a complementary desire to act unimpeded and maintain freedom (negative face). Similar to speech acts, utterances can perform so-called face acts which can either raise or threaten the positive or negative face of the speaker or hearer. We begin by using an existing corpus to train a model which classifies face acts, achieving a new SoTA in the process. We then observe that every face act has an underlying intention that motivates it and perform additional experiments integrating dialog act annotations to provide these intentions by proxy. Our analysis finds that dialog acts improve performance on face act detection for minority classes and points to a close relationship between aspects of face and intent.
\\ \newline 
}

\begin{document}

\maketitleabstract

% \begin{abstract}
% \citet{brown-levinson-1987-politeness} introduce a highly influential framework for understanding politeness through the notion of face. In this theory agents attend to their desire to be appreciated and wanted, which they call positive face, and a complementary desire to act unimpeded and maintain freedom, which they call negative face.
% Utterances can either raise or threaten the positive or negative face of the speaker or addressee.
% \citet{dutt-etal-2020-keeping} present the novel task of automatically labeling so-called face acts following this theory. In this paper we continue this line of study by employing a sequence-to-sequence approach, achieving SoTA performance with an F-measure of 0.73. We then observe that every face act has an underlying intention which motivates it and perform additional experiments integrating dialogue act annotations to provide these intentions by proxy. The results are used to conduct a set of analyses which point to a close relationship between aspects of politeness and intent.
% \end{abstract}

\section{Introduction}
\label{sec:intro}

% Proposed structure of intro:
% - Intro to Brown \& Levinson (good short intro in abstract, ok to reuse partially), also CMU corpus

% \citet{brown-levinson-1987-politeness} introduce a highly influential framework for understanding politeness through the notion of face. In this theory agents attend to their desire to be appreciated and wanted, which they call positive face, and a complementary desire to act unimpeded and maintain freedom, which they call negative face. Utterances can either raise or threaten the positive or negative face of the speaker or addressee.   

\citet{brown-levinson-1987-politeness} introduce an influential theory of politeness based on the concept of ``face'', which they claim to be culturally universal. In this theory, face -- i.e., the public image one seeks to claim -- is a two-sided coin. Agents attend to their desire to have their wants appreciated, which the theory calls positive face, as well as a complementary desire to act unimpeded and maintain freedom, which the theory calls negative face. The face of every agent is ensnared with that of every other agent -- agents cannot have their desires appreciated if they cannot appreciate the desires of others. As a result, utterances can raise (+) or threaten (-) the positive (Pos) or negative (Neg) face of the speaker (S) or hearer (H).

\begin{table*}[!ht]
\centering
\resizebox{0.9\linewidth}{!}{
\begin{tabular}{lll}
\toprule
\bfseries Face Act & \bfseries Interpretation & \bfseries Example Discourse Goals \\
\midrule
\HNegT & Imposition & Requests, commands, questions, offers, promises, ... \\ 
\HPosT & Disagreement & Criticism, insults, disapproval, ... \\
\HNegR & Permissiveness & Granting permission, making exceptions, ... \\
\HPosR & Agreement & Seeking common ground, group cohesion, ... \\
\midrule
\SNegT & Indebtedness & Thanking, accepting offers or thanks, commitments, ... \\
\SPosT & Apologies & Confessions, embarrassment, ... \\
\SNegR & Autonomy & Refusing requests, asserting freedoms, ... \\
\SPosR & Confidence & Self-promotion, signaling virtue, ... \\
\bottomrule
\end{tabular}
}
\caption{Face acts with a short label, which serves as an general interpretation of the face act, and some examples of their related discourse goals.}
\label{tab:faceacts}
\end{table*}

% XXX: NEW
For example, \emph{pass the salt} would be labeled \HNegT as it imposes on the hearer's freedom by requesting their action, while \emph{that was my fault}, a self-critique, would be considered \SPosT. A summary of possible face acts is shown in Table \ref{tab:faceacts}.
% XXX: NEW

% - Make observation that politeness is not a property of LANGAUGE used, but of communicative intent: we threaten or raise because of what we want to do with language, and then we mitigate the effect using language.  Therefore: this paper investigates how modeling intent in dialog helps FTA detection.

A face threat or face raising is not a property of particular linguistic choices, but of communicative intent. If I want to request information from you, then I necessarily need to threaten your negative face: if you recognize the speech act (and thus it is successful), I will oblige you to answer, and therefore I will restrict your choice of actions.  In \citet{brown-levinson-1987-politeness}'s theory, discourse participants choose among various strategies for minimizing threats to face. These strategies are linguistic strategies (for example, using hedges), and the choice of strategy depends on many factors such as  the discourse situation (who is talking to whom under what circumstances) and cultural conventions.
% These strategies are linguistic strategies (for example, using hedges), and the choice of strategy depends on many factors, in particular on cultural conventions and on the discourse situation (who is talking to whom under what circumstances).

Work related to natural language processing has concentrated on studying linguistic manifestations of politeness \cite{walker-etal-1997-improvising,danescu-niculescu-mizil-etal-2013-computational}, while largely disregarding the notion of face act (FA).  The major exception is the seminal work of \citet{dutt-etal-2020-keeping}, who annotate a corpus of written dialogs for face acts, and then develop a system for predicting face acts for dialog turns.  In this paper, we build on this work.  Our goal is not primarily to improve on the tagging results (which we do), but to understand better how face acts interact with discourse intention, and to pave the way for a new phase of work on face acts. 
We believe face acts, as conceived of by \citet{brown-levinson-1987-politeness}, are not a peripheral aspect of NLP, but can serve a crucial role in improving both interactive NLP and our understanding of how we humans use language.  
There are two reasons to make such a strong claim.  First, in NLP it has been observed that, despite their great success in many engineering problems, large language models (LLMs) are typically pre-trained entirely on sequences of words and do not model intention \cite{bender-etal-2021-parrots,bender-koller-2020-climbing}.  However, communicative intention and intention recognition is the fundamental mechanism of communication.
% \owenshort{Is that what Bender says?  Need to check -- yes and she cite brennen and clark for it!!!}
Second, while much progress has been made (again partially thanks to LLMs) in processing multiple languages, there has not been much work in NLP that addresses the culture-specific ways in which language is used in context.  We believe the study of face acts can address both issues.  This paper is a first step in the direction of a larger research project.

% - Dialog acts as models of intent, lots of existing work
%\noindent
%This paper makes two main contributions: 

% \begin{enumerate}
% \item
% We provide a relatively straightforward application of generative neural techniques to the FA tagging problem, and we obtain a new state-of-the-art (from 69\% to 73\% F-measure).
% \item
% In a second set of experiments, we train using information from dialog act taggers, and we show in an extensive analysis how this helps for specific FAs, which are hard to tag without such information.
% \end{enumerate}

The principal contribution of this paper lies in a set of experiments in which we train FA taggers using information from dialog act taggers.  We show in an extensive analysis how this helps for specific FAs, which are hard to tag without such information.  Furthermore, we provide a relatively straightforward application of generative neural techniques to the FA tagging problem, and we obtain a new state-of-the-art (increasing the state-of-the-art from 69\% to 73\% F-measure).

% \noindent
This paper is structured as follows. We start with a review of relevant literature (\S\ref{sec:related}) and an outline of our approach to modeling and evaluation (\S\ref{sec:approach}). We then discuss our experiments using only face act information (\S\ref{sec:fos}) as well as follow-up experiments which integrate intention through the use of dialog acts (\S\ref{sec:adding-intent}). We then conduct an extensive error analysis of all system variants (\S\ref{sec:error-analysis}) and report our conclusions along with a discussion of future work for this research program (\S\ref{sec:conclusion}).

\section{Related Work}

\label{sec:related}

\citet{brown-levinson-1987-politeness} provide a theory of politeness which has been fundamental to work in various fields concerned with how language is used. We have provided a brief summary in Section~\ref{sec:intro}.   Curiously, in NLP there has not been much work building explicitly on \citep{brown-levinson-1987-politeness}. \citet{danescu-niculescu-mizil-etal-2013-computational} concentrate on one type of face-threatening act (FTA), namely the negative face-threatening act of a request, and investigate the strategies used for doing this FTA.  To do this, they use crowd sourcing to rate the requests on a politeness scale. They then develop a model which predicts the politeness of these requests and use it to study how this affects the interactions between users on Wikipedia and StackExchange.
% They then develop a politeness predictor which they use to study the use of politeness in two large corpora in which hierarchies exist and show how these hierarchies interact with politeness.

The face acts (FAs) themselves are the object of \citet{dutt-etal-2020-keeping}. In addition to developing a data set annotated with FAs, they present a FA classifier based on a neural architecture they devise on top of BERT, which achieves 69\% F-measure (60\% macro). As the data involves participants convincing others to donate to a charity, they also use this corpus to investigate the relationship between face acts and persuasion by predicting if a participant chose to donate. We use this data set in our work on FA tagging.
% In our work, we use this data set, and employ a fairly simple transformer architecture with fine-tuning to achieve a new state-of-the-art F1 for FA tagging of 73\%.

A salient aspect of the work of \citet{brown-levinson-1987-politeness} is that they situate the notion of politeness within a larger theory of rational interaction, as outlined by \citet{grice:1975}.  One consequence of this is that successful communication requires the recognition of intent: a speaker's request cannot threaten the hearer's negative face if the hearer does not recognize the speech act as a request.  There is a large body of work on speech acts and intent, starting with \citet{austin:1962}.  We do not provide a summary of all relevant work here, but the notion of speech act was extended in NLP as a dialog act, and given several fine-grained inventories that go well beyond the initial high-level distinctions of speech acts \cite{anderson1991hcrc,core-and-allen-1997-coding,S*:2000}.  The corresponding classification task, dialog act tagging, is a crucial component in creating dialog systems, as it allows for a simple way of modeling the user's communicative intent through text classification. In this paper, we do not make contributions to dialog act tagging, but we use existing work.

\section{Approach} \label{sec:approach}

In this section we outline the data, modeling techniques, and evaluation measures used throughout the paper.

\subsection{Face Act Data}

As discussed in Section~\ref{sec:related}, we use the face data set developed by \citet{dutt-etal-2020-keeping} for our experiments. \citet{wang-etal-2019-persuasion} introduce a corpus of dyadic, persuasion-oriented conversations sourced from an online task where Amazon Mechanical Turk workers must convince their addressee to donate part of their task earnings to a charity, Save the Children. The conversations are carried out through a chat interface with one worker acting as the persuader (ER) and the other as the persuadee (EE). The participants were informed that the dialog must last at least 10 turns and that their reward is not penalized should they fail to convince their partner to donate. \citet{dutt-etal-2020-keeping} augment conversations from this corpus with eight face act annotations (see Table~\ref{tab:fos-results}) based on \citet{brown-levinson-1987-politeness}. 
% === NEW
They take some small departures from politeness theory in their annotation. Most notably, thanking is annotated as \HPosR rather than \SNegT and
% === NEW
\Other is used to indicate that no face act is present. The authors selected the corpus as their starting point for two main reasons. First, the goal-oriented nature of the conversations incentivizes face-threatening acts, which are typically avoided unless necessary. Second, each participant is on equal ground which helps mitigates potentially confounding issues, like power distance. % Either party is equally free to choose non-compliance or even leave the conversation altogether. 

% There is notably, no instance in this corpus of speaker positive face threatening (\SNegT) which the authors attribute to genre.
% \owenshort{I stole two paragraphs for related work.  Add here some info about size of corpus.  Maybe point to table with distribution of labels as well (which includes results on classification, so it comes later)?}
% The corpus contains eight labels (see Table~\ref{tab:fos-results}) with one to indicate no face act present (\Other).
% Add info on what the inventory of tags is here.  Should have tagset of size 8.

It is possible for a single utterance to perform multiple face acts at once. For example, \emph{Just stick to what you know} could be seen as both \HNegT, since it is issuing a command, and \HPosT, as it entails the critique that they did not know what they were doing. However, \citet{dutt-etal-2020-keeping} observed multi-labeled acts in only 2\% of their data leading them to simplify the the problem to a single label per utterance. In the event of a multi-label annotation, they select one randomly. The resulting data set contains 10,716 turns averaging 10 words (or 51 characters) in length across 296 unique conversations. The label distribution (see Table~\ref{tab:fos-results}) is highly imbalanced with vanishingly rare labels like \tsc{SPos+} appearing only 12 times and labels like \tsc{SNeg-} actually vanishing.
%, the latter of which is attributed to the genre.

\subsection{Modeling}
We model face act tagging as a text classification task. Given a sequence of $n$ tokens $S = [t_1, t_2, \dots, t_n]$, we wish to assign a label $y \in Y$ where $Y$ represents the set containing the 8 possible face acts. Recently, similar classification tasks including sentiment analysis \citep{zhang-etal-2021-towards-generative} and event factuality prediction \citep{murzaku-etal-2022-examining} have achieved state-of-the-art results training sequence-to-sequence models. We adopt this approach and utilize Flan-T5-base \citep{chung-etal-2022-scaling} for fine-tuning which allows us to re-frame the problem as a generation task with limited model-specific requirements.\footnote{Our choice of Flan-T5 was informed by a preliminary set of small experiments in which a variety of pre-trained models were were examined on single seed runs.} % \owenshort{Could pull footnote into main text}

\subsection{Data Representation} \label{sec:data-representation}

While generative approaches unify many aspects of the model design, they present challenges when it comes to determining effective input and output representations.
We provide the models an input which contains an utterance prefixed with ER for persuaders or EE for persuadees along with up to two previous turns of dialog as context.
Each turn is separated by a newline character which we treat as a special token when tokenizing.

{\addtolength\leftmargini{-0.1in}
\begin{quote}
\small
\ttfamily
\noindent
\hspace{-1em}[\tbf{Input}] \\
\noindent
ER: Are you interested in donating? \\
EE: Possibly, I'm not sure. \\
EE: I don't even know what the charity is.

\hspace{-1em}[\tbf{Output}] \\
sneg+
\end{quote}}

% \begin{minted}{text}
% [Input]
%     ER: Are you interested in donating?
%     EE: Possibly, I'm not sure.
%     EE: I don't even know what the charity is.
% [Output]
%     sneg+
% \end{minted}

\noindent
The target output is a string containing the correct label for the final utterance of the input text, in this case \SNegR since the speaker is asserting their freedom of action. In preliminary experiments we found context to be a critical factor with a size of two, for a total of three utterances, performing best. As there are no previous turns for the first two turns in each dialog, those examples are provided in a similar format containing only one or no lines of context. % \owenshort{Reformulated -- check end of thsi sentence}

\subsection{Evaluation} \label{sec:approach-eval}

% XXX: Revise this idea of advocating for macro removed.
We evaluate model performance using F-measure for each of the eight represented classes as well as micro and macro F-measure aggregated over all labels. Since each utterance is assigned exactly one label, micro F-measure is the same as accuracy. We observe that the extreme rarity of \SPosR (12 occurrences) contributes to high variance in macro F1 and, as a result, advocate focusing on micro F1 and macro F1 with this label removed as the primary high-level metrics for this task. So as to maintain comparability with previous work, we report values for macro F-measure computed with all represented labels.
% \owenshort{So we advocate for not using it but we do use it?}
Our experiments are performed using five-fold cross-validation on the same splits which \citet{dutt-etal-2020-keeping} report their findings on.\footnote{\url{https://github.com/xinru1414/Face_Act}} We identified some conversations which were duplicated across folds and keep only the first appearance of these entries when performing evaluation. The evaluation metrics are averaged across all five folds.

Ideally, the output generated by the model would contain only sequences in our label set, but there were a small number of cases in which malformed output was produced. We repair these labels using a methodology inspired by \cite{zhang-etal-2021-towards-generative}. Invalid output sequences are compared with all possible labels using edit-distance and the closest match is used. In the event of a tie, the most frequent label in the training data is chosen. 

\subsection{Replication}
% All of the code written, data sets prepared, and experimental observations made in the course of this research will be made available on GitHub. % We perform two groups of experiments.\todo{insert summary here}
All of the code written, data sets prepared, and experimental observations made in the course of this research are available on \href{https://github.com/cogstates/2024-lrec-coling-faceacts}{GitHub}.\footnote{\url{https://github.com/cogstates/2024-lrec-coling-faceacts}}

\section{Face-Only System} \label{sec:fos}

We begin by training the model with no additional information containing intentions. We will refer to this configuration as the \FOS (Face-Only System).

\paragraph{Methodology} We fine-tune Flan-T5-base on each of the five cross-validation folds with a batch size of 32 and two gradient accumulation steps. The AdamW optimizer is configured with a learning rate of 3e-4, weight decay of 0.01, and epsilon of 1e-8. As the cross-validation preparation does not contain a development set, early stopping is configured using micro F1 on the training data set with a patience of 3. In general, fine-tuning completed after roughly 15 to 20 epochs. These parameters were arrived at through a small run of hyperparameter tuning experiments. When predicting, generation is performed with a beam size of one.

\begin{table}[!t]
\centering
\resizebox{\linewidth}{!}{
\begin{tabular}{lc|ccc|l}
\toprule
& \bfseries F1 & \bfseries F1 & \bfseries Prec. & \bfseries Recall & \bfseries Count \\
\midrule
\bfseries Macro & 0.60 & 0.63 & 0.63 & 0.63 & - \\
\bfseries Micro & 0.69 & 0.73 & 0.73 & 0.73 & - \\
\midrule
\bfseries \Other & - & 0.75 & 0.76 & 0.73 & 4,300 \\
\bfseries \HPosR & - & 0.75 & 0.72 & 0.77 & 2,844 \\
\bfseries \SPosR & - & 0.74 & 0.74 & 0.75 & 1,589 \\
\bfseries \HNegT & - & 0.74 & 0.71 & 0.76 & 1,073 \\
\cdashlinelr{1-6}
\bfseries \HPosT & - & 0.55 & 0.61 & 0.51 & 334 \\
\bfseries \HNegR & - & 0.44 & 0.47 & 0.41 & 305 \\
\bfseries \SNegR & - & 0.57 & 0.61 & 0.53 & 259 \\
\cdashlinelr{1-6}
\bfseries \SPosT & - & 0.47 & 0.39 & 0.58 & 12 \\
\midrule
\multicolumn{2}{r}{\citet{dutt-etal-2020-keeping}} & \multicolumn{3}{|c|}{\FOS} \\
\bottomrule
\end{tabular}}
\caption{Performance of our \FOS against the previous state-of-the-art BERT HiGRU-f model and label count.}
\label{tab:fos-results}
\end{table}

% \begin{table}[t]
% \begin{center}
% % \begin{tabular}{|l|r|r|}
% % \hline
% % Label & Count & F-measure \\
% % \hline
% % \Other  &  4300 &  74.5\%\\
% % \HPosR  &  2844 &  74.6\%\\
% % \SPosR  &  1589 & 73.9\%\\
% % \HNegT  &  1073 & 73.5\%\\
% % \hline
% % %MIN WITH PROBLEMS
% % \HPosT  &   334  & 55.4\%\\
% % \HNegR  &   305  & 44.1\%\\
% % \SNegR  &   259  & 55.3\%\\
% % \hline
% % %IGNORE
% % \SPosT  &    12 & 50.8\%\\\hline
% % \end{tabular}
% \begin{tabular}{lrr}
% \toprule
% \bfseries Label & \bfseries Count & \bfseries F-measure \\
% \midrule
% \Other & 4,300 & 0.75 \\
% \HPosR & 2,844 & 0.75 \\
% \SPosR & 1,589 & 0.74 \\
% \HNegT & 1,073 & 0.74 \\
% \HPosT & 334 & 0.55 \\
% \HNegR & 305 & 0.44 \\
% \SNegR & 259 & 0.57 \\
% \SPosT & 12 & 0.47 \\
% \bottomrule
% \end{tabular}
% \caption{Distribution of labels and F-measure for \FOS.}
% \label{tab:result-dist}
% \end{center}
% \end{table}

\paragraph{Results} \label{sec:fos-results} Evaluation metrics, averaged across all folds, for the \FOS are summarized in Table~\ref{tab:fos-results}.
% \owenshort{Should I add the counts into this table? Basically consolidating it with Table 5?  Also, reorder by frequency?}
This relatively straight-forward approach achieves a macro F1 of 0.63 and micro F1 of 0.73, improving on the previous state-of-the-art by 3 and 4 points, respectively. 
% Unsurprisingly, micro F-measure\owenshort{You mean simply F1, right?  Say "Among the labels, F1 correlates..."} correlates strongly ($r = 0.77$) with the number of examples found in the data. 
Among the labels, F1 correlates strongly ($r = 0.77$) with the number of examples found in the data.
In other words, the minority classes are where this model struggles to find signal.

\begin{table*}[!t]
\centering
\scalebox{1}{
\resizebox{\linewidth}{!}{
\begin{tabular}{lccc|ccc|ccc|ccc|ccc}
\toprule
\multicolumn{1}{c|}{} & \multicolumn{3}{c|}{\bfseries \SPosR} & \multicolumn{3}{c|}{\bfseries \HPosT} & \multicolumn{3}{c|}{\bfseries \SNegR} & \multicolumn{3}{c|}{\bfseries \HNegR} & \multicolumn{3}{c}{\bfseries \HNegT} \\
\multicolumn{1}{c|}{} & \bfseries F1\sddag & \bfseries Prec.\sdag & \bfseries Recall & \bfseries F1 & \bfseries Prec. & \bfseries Recall & \bfseries F1 & \bfseries Prec. & \bfseries Recall\sdag & \bfseries F1\sddag & \bfseries Prec.\sdag & \bfseries Recall\sddag & \bfseries F1 & \bfseries Prec. & \bfseries Recall\sdag \\
\midrule
\bfseries \FOS & {\cellcolor[HTML]{023858}} \color[HTML]{F1F1F1} 0.74 & {\cellcolor[HTML]{023B5D}} \color[HTML]{F1F1F1} 0.74 & {\cellcolor[HTML]{023858}} \color[HTML]{F1F1F1} 0.75 & {\cellcolor[HTML]{023858}} \color[HTML]{F1F1F1} 0.55 & {\cellcolor[HTML]{023858}} \color[HTML]{F1F1F1} 0.61 & {\cellcolor[HTML]{97B7D7}} \color[HTML]{000000} 0.51 & {\cellcolor[HTML]{023858}} \color[HTML]{F1F1F1} 0.57 & {\cellcolor[HTML]{03466E}} \color[HTML]{F1F1F1} 0.61 & {\cellcolor[HTML]{023858}} \color[HTML]{F1F1F1} 0.53 & {\cellcolor[HTML]{C2CBE2}} \color[HTML]{000000} 0.44 & {\cellcolor[HTML]{81AED2}} \color[HTML]{F1F1F1} 0.47 & {\cellcolor[HTML]{D9D8EA}} \color[HTML]{000000} 0.41 & {\cellcolor[HTML]{023858}} \color[HTML]{F1F1F1} 0.74 & {\cellcolor[HTML]{03446A}} \color[HTML]{F1F1F1} 0.71 & {\cellcolor[HTML]{62A2CB}} \color[HTML]{F1F1F1} 0.76 \\
\bfseries \TAS & {\cellcolor[HTML]{F8F1F8}} \color[HTML]{000000} 0.70 & {\cellcolor[HTML]{FFF7FB}} \color[HTML]{000000} 0.68 & {\cellcolor[HTML]{80AED2}} \color[HTML]{F1F1F1} 0.72 & {\cellcolor[HTML]{2D8ABD}} \color[HTML]{F1F1F1} 0.53 & {\cellcolor[HTML]{3991C1}} \color[HTML]{F1F1F1} 0.56 & {\cellcolor[HTML]{ADC1DD}} \color[HTML]{000000} 0.51 & {\cellcolor[HTML]{FFF7FB}} \color[HTML]{000000} 0.48 & {\cellcolor[HTML]{FFF7FB}} \color[HTML]{000000} 0.55 & {\cellcolor[HTML]{FFF7FB}} \color[HTML]{000000} 0.43 & {\cellcolor[HTML]{045A8D}} \color[HTML]{F1F1F1} 0.49 & {\cellcolor[HTML]{FFF7FB}} \color[HTML]{000000} 0.44 & {\cellcolor[HTML]{023858}} \color[HTML]{F1F1F1} 0.56 & {\cellcolor[HTML]{FFF7FB}} \color[HTML]{000000} 0.72 & {\cellcolor[HTML]{60A1CA}} \color[HTML]{F1F1F1} 0.69 & {\cellcolor[HTML]{B9C6E0}} \color[HTML]{000000} 0.75 \\
\bfseries \TAM & {\cellcolor[HTML]{FFF7FB}} \color[HTML]{000000} 0.70 & {\cellcolor[HTML]{A9BFDC}} \color[HTML]{000000} 0.70 & {\cellcolor[HTML]{FFF7FB}} \color[HTML]{000000} 0.69 & {\cellcolor[HTML]{4496C3}} \color[HTML]{F1F1F1} 0.53 & {\cellcolor[HTML]{056EAD}} \color[HTML]{F1F1F1} 0.58 & {\cellcolor[HTML]{FFF7FB}} \color[HTML]{000000} 0.49 & {\cellcolor[HTML]{A5BDDB}} \color[HTML]{000000} 0.51 & {\cellcolor[HTML]{2685BB}} \color[HTML]{F1F1F1} 0.59 & {\cellcolor[HTML]{D2D3E7}} \color[HTML]{000000} 0.45 & {\cellcolor[HTML]{023858}} \color[HTML]{F1F1F1} 0.51 & {\cellcolor[HTML]{8BB2D4}} \color[HTML]{000000} 0.47 & {\cellcolor[HTML]{023C5F}} \color[HTML]{F1F1F1} 0.55 & {\cellcolor[HTML]{D2D3E7}} \color[HTML]{000000} 0.72 & {\cellcolor[HTML]{023858}} \color[HTML]{F1F1F1} 0.71 & {\cellcolor[HTML]{FFF7FB}} \color[HTML]{000000} 0.73 \\
\bfseries \MTS & {\cellcolor[HTML]{65A3CB}} \color[HTML]{F1F1F1} 0.72 & {\cellcolor[HTML]{023858}} \color[HTML]{F1F1F1} 0.74 & {\cellcolor[HTML]{D5D5E8}} \color[HTML]{000000} 0.70 & {\cellcolor[HTML]{045687}} \color[HTML]{F1F1F1} 0.55 & {\cellcolor[HTML]{529BC7}} \color[HTML]{F1F1F1} 0.56 & {\cellcolor[HTML]{023858}} \color[HTML]{F1F1F1} 0.54 & {\cellcolor[HTML]{0A73B2}} \color[HTML]{F1F1F1} 0.54 & {\cellcolor[HTML]{CACEE5}} \color[HTML]{000000} 0.57 & {\cellcolor[HTML]{034369}} \color[HTML]{F1F1F1} 0.52 & {\cellcolor[HTML]{FFF7FB}} \color[HTML]{000000} 0.41 & {\cellcolor[HTML]{D5D5E8}} \color[HTML]{000000} 0.46 & {\cellcolor[HTML]{FFF7FB}} \color[HTML]{000000} 0.37 & {\cellcolor[HTML]{FDF5FA}} \color[HTML]{000000} 0.72 & {\cellcolor[HTML]{FFF7FB}} \color[HTML]{000000} 0.65 & {\cellcolor[HTML]{023858}} \color[HTML]{F1F1F1} 0.79 \\
\bfseries \MTM & {\cellcolor[HTML]{6FA7CE}} \color[HTML]{F1F1F1} 0.72 & {\cellcolor[HTML]{569DC8}} \color[HTML]{F1F1F1} 0.71 & {\cellcolor[HTML]{328DBF}} \color[HTML]{F1F1F1} 0.73 & {\cellcolor[HTML]{FFF7FB}} \color[HTML]{000000} 0.49 & {\cellcolor[HTML]{FFF7FB}} \color[HTML]{000000} 0.49 & {\cellcolor[HTML]{D3D4E7}} \color[HTML]{000000} 0.50 & {\cellcolor[HTML]{3790C0}} \color[HTML]{F1F1F1} 0.53 & {\cellcolor[HTML]{023858}} \color[HTML]{F1F1F1} 0.62 & {\cellcolor[HTML]{8CB3D5}} \color[HTML]{000000} 0.47 & {\cellcolor[HTML]{CCCFE5}} \color[HTML]{000000} 0.43 & {\cellcolor[HTML]{023858}} \color[HTML]{F1F1F1} 0.50 & {\cellcolor[HTML]{F2ECF5}} \color[HTML]{000000} 0.39 & {\cellcolor[HTML]{045D92}} \color[HTML]{F1F1F1} 0.73 & {\cellcolor[HTML]{187CB6}} \color[HTML]{F1F1F1} 0.70 & {\cellcolor[HTML]{0F76B3}} \color[HTML]{F1F1F1} 0.78 \\
\bottomrule
\end{tabular}

}}
\caption{Detailed evaluation results for all experiments on the five least common labels, excluding \SPosT. Significant differences using the Friedman rank sum test are marked with \dag\ and \ddag\ for $\alpha = 0.10$ and $0.05$ respectively.}
\label{tab:da-results-min}
\end{table*}

\begin{table}[tb]
\centering
\resizebox{\linewidth}{!}{
\begin{tabular}{l|cccccc}
\toprule
& \multicolumn{2}{c}{\bfseries F1} & \multicolumn{2}{c}{\bfseries Precision} & \multicolumn{2}{c}{\bfseries Recall} \\
& \bfseries Micro & \bfseries Macro & \bfseries Micro & \bfseries Macro & \bfseries Micro & \bfseries Macro \\
\midrule
\bfseries \FOS & {\cellcolor[HTML]{023858}} \color[HTML]{F1F1F1} 0.73 & {\cellcolor[HTML]{023858}} \color[HTML]{F1F1F1} 0.63 & {\cellcolor[HTML]{023858}} \color[HTML]{F1F1F1} 0.73 & {\cellcolor[HTML]{023858}} \color[HTML]{F1F1F1} 0.63 & {\cellcolor[HTML]{023858}} \color[HTML]{F1F1F1} 0.73 & {\cellcolor[HTML]{023F64}} \color[HTML]{F1F1F1} 0.63 \\
\bfseries \TAS & {\cellcolor[HTML]{D7D6E9}} \color[HTML]{000000} 0.70 & {\cellcolor[HTML]{2987BC}} \color[HTML]{F1F1F1} 0.61 & {\cellcolor[HTML]{FFF7FB}} \color[HTML]{000000} 0.70 & {\cellcolor[HTML]{ADC1DD}} \color[HTML]{000000} 0.59 & {\cellcolor[HTML]{FFF7FB}} \color[HTML]{000000} 0.70 & {\cellcolor[HTML]{023858}} \color[HTML]{F1F1F1} 0.63 \\
\bfseries \TAM & {\cellcolor[HTML]{ACC0DD}} \color[HTML]{000000} 0.70 & {\cellcolor[HTML]{79ABD0}} \color[HTML]{F1F1F1} 0.60 & {\cellcolor[HTML]{E1DFED}} \color[HTML]{000000} 0.70 & {\cellcolor[HTML]{7BACD1}} \color[HTML]{F1F1F1} 0.60 & {\cellcolor[HTML]{E1DFED}} \color[HTML]{000000} 0.70 & {\cellcolor[HTML]{7BACD1}} \color[HTML]{F1F1F1} 0.60 \\
\bfseries \MTS & {\cellcolor[HTML]{C8CDE4}} \color[HTML]{000000} 0.70 & {\cellcolor[HTML]{FFF7FB}} \color[HTML]{000000} 0.57 & {\cellcolor[HTML]{F5EEF6}} \color[HTML]{000000} 0.70 & {\cellcolor[HTML]{FFF7FB}} \color[HTML]{000000} 0.58 & {\cellcolor[HTML]{F5EEF6}} \color[HTML]{000000} 0.70 & {\cellcolor[HTML]{FFF7FB}} \color[HTML]{000000} 0.57 \\
\bfseries \MTM & {\cellcolor[HTML]{2F8BBE}} \color[HTML]{F1F1F1} 0.71 & {\cellcolor[HTML]{549CC7}} \color[HTML]{F1F1F1} 0.60 & {\cellcolor[HTML]{60A1CA}} \color[HTML]{F1F1F1} 0.71 & {\cellcolor[HTML]{1379B5}} \color[HTML]{F1F1F1} 0.61 & {\cellcolor[HTML]{60A1CA}} \color[HTML]{F1F1F1} 0.71 & {\cellcolor[HTML]{96B6D7}} \color[HTML]{000000} 0.59 \\
\bfseries \hyperlink{cite.dutt-etal-2020-keeping}{Dutt et al.} & {\cellcolor[HTML]{FFF7FB}} \color[HTML]{000000} 0.69 & {\cellcolor[HTML]{5C9FC9}} \color[HTML]{F1F1F1} 0.60 & {\cellcolor[HTML]{FFFFFF}} \color[HTML]{000000} - & {\cellcolor[HTML]{FFFFFF}} \color[HTML]{000000} - & {\cellcolor[HTML]{FFFFFF}} \color[HTML]{000000} - & {\cellcolor[HTML]{FFFFFF}} \color[HTML]{000000} - \\
\bottomrule
\end{tabular}}
\caption{Summary of model results.}
\label{tab:da-results}
\end{table}

\section{Adding Intention} \label{sec:adding-intent}

We observe that every face act has an underlying intention which motivates it. A rational agent would not risk 
%their own face
reprisals that result from 
% \owenshort{Or 
threatening their interlocutor's face
%, due to fear of reprisal?} 
unless (1) they have a goal or intention which necessitates the face act or (2) other cultural factors such as power-distance protect them from such reprisals. As (2) is not the case in our corpus, we have reason to believe that providing the model with explicit information regarding agent intentions may improve performance. % We experiment with two methods of integrating this information into the model using dialog acts.

% FROM REBUTTALS
% We agree that dialogue acts fail to represent hierarchical (specifically, higher-level) intentions and leave exploring this issue to future work. We note that while DA tagging is a well established NLP task, hierarchical intention recognition is not

\subsection{Data}
We represent information regarding intent using two well-known dialog act corpora with varying levels of granularity in distinguishing actions. % \todo{expand why this is a good idea}

\paragraph{Meeting Recorder Dialog Act Corpus} MRDA \citep{shriberg-etal-2004-icsi} consists of transcripts from 75 in-person meetings that are generally research oriented in nature, annotated for dialog acts using a tag set adapted from DAMSL \citep{core-and-allen-1997-coding}. It contains 108,202 utterances with a mean utterance length of 8 words.
The annotations are provided in three levels of granularity. The ``basic'' level contains 6 tags, the ``general'' level contains 12 tags, and the full label set contains 52 tags. We utilize the basic tag set for our experiments.
% \owenshort{Why that citation -- they do teh same?  Maybe say so?} % \cite{ang-etal-2005-automatic}

\paragraph{Switchboard Dialog Act Corpus} SWDA is a collection of short phone conversations in which callers are matched with a partner to discuss some provided general-interest topic \cite{stolcke-etal-2000-dialogue}. It contains roughly 180,000 utterances with a mean length of 9.6 words
% \owenshort{words i assume?  maybe say}
across 1,155 unique conversations. They use the DAMSL dialog act tag set to annotate the transcripts with 41 dialog act labels.

% TODO: Clarify this.
% > A: In section 5.2, "use the dialog act tagging [...] two new preparations" still not clear to me. Can the authors provide more details with regard to how their text-augmented models work?
%
% The text-augmented models work very similarly to the FoS system with the only difference being that we modified (“augmented”) the input text used in training by including the dialog act model predictions in parenthesis at the end of each line of dialog in the input text. The two preparations refer to one using the SWDA tags and another using the MRDA tags. An example of this augmented input is shown in section 5.2. The idea is that the model might pick up on these patterns in DA labels during training. We will look at revising the writing in this section to be more clear.

\subsection{Methodology} \label{sec:methodology-da}
% \owenshort{Rewrote this a bit for clarity, please check and fill in info about MRDA}
We experiment with two methods of integrating dialog act information into the model. In the first method, we augment the face act corpus with explicit dialog act tags in the text. For both MRDA and SWDA, we use the dialog act tagging system of \citet{he-etal-2021-speaker-turn} to automatically augment the face act corpus and produce two new preparations of data for training.
%We repeat this process for MRDA which produces two new preparations of the face act data. 
We will refer to the resulting text-augmented models as \TAM and \TAS, respectively.
Returning to the example in Section \ref{sec:data-representation}, the augmented input using SWDA tags is shown below.  

{\addtolength\leftmargini{-0.1in}
\begin{quote}
\small
\ttfamily
\noindent
\hspace{-1em}[\tbf{Input}] \\
\noindent
ER: Are you interested in donating? (Yes-No-Question) \\ % (Wh-Question) \\ % (Question)
EE: Possibly, I'm not sure. (Hedge) \\ % (Statement)
EE: I don't even know what the charity is. (Statement-non-opinion) % (Statement)

\hspace{-1em}[\tbf{Output}] \\
sneg+
\end{quote}}

% \begin{minted}{text}
% [Input]
%     ER: Are you interested in donating? (Yes-No-Question)
%     EE: Possibly, I'm not sure. (Hedge)
%     EE: I don't even know what the charity is. (Statement-non-opinion)
% [Output]
%     sneg+
% \end{minted}

\noindent
In the second method, we incorporate the dialog act data through traditional multi-task learning. As the dialog act data sets contain far more examples than the face act data set, we randomly sample 10\% of conversations from both SWDA and MRDA. This results in training regimens with roughly 1:1 and 2:1 ratios of dialog acts to face acts, respectively. 
% To indicate the desired task, each example is prefixed accordingly. % -- BEFORE REVISION
To indicate the desired task, each example input is prefixed accordingly with the task name (\emph{dialog acts} or \emph{face acts}) followed by a colon and new line.
% \owenshort{If we have space, maybe list the prefixes?}
The data for both tasks is provided with the same three total turns of context as used for the \FOS. We will refer to the resulting models as \MTM and \MTS.

%The training configuration, cross-validation folds, and hardware used for building the \FOS are held constant in these experiments.
These experiments use the same training configuration, cross-validation folds, and hardware as we used in the case of \FOS. % \owenshort{I edited for clarity -- still correct?}

\begin{figure}[!b]
    \centering
    \resizebox{\linewidth}{!}{
    \includegraphics[width=\textwidth]{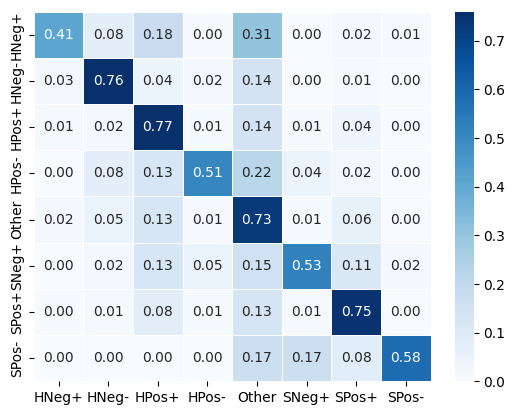}
    }
    \caption{Confusion matrix for the \FOS; the numbers show the fraction of time the tag on the x-axis is predicted instead of the gold label on the y-axis; the numbers in each row add up to 1.}
    \label{fig:basic-confusion}
\end{figure}

\begin{table*}[htb]
\centering
\resizebox{\linewidth}{!}{
\begin{tabular}{lccccccccc}
\toprule
& \bfseries \HNegR & \bfseries \HNegT & \bfseries \HPosR & \bfseries \HPosT & \bfseries \Other & \bfseries \SNegR & \bfseries \SPosR & \bfseries \SPosT & \bfseries Total \\
\midrule
\bfseries Both Happening (Same Part) & {\cellcolor[HTML]{8BB2D4}} \color[HTML]{000000} 5 & {\cellcolor[HTML]{E6E2EF}} \color[HTML]{000000} 2 & {\cellcolor[HTML]{DBDAEB}} \color[HTML]{000000} 3 & {\cellcolor[HTML]{023858}} \color[HTML]{F1F1F1} 7 & {\cellcolor[HTML]{FFF7FB}} \color[HTML]{000000} 0 & {\cellcolor[HTML]{04598C}} \color[HTML]{F1F1F1} 7 & {\cellcolor[HTML]{2685BB}} \color[HTML]{F1F1F1} 5 & {\cellcolor[HTML]{FFF7FB}} \color[HTML]{000000} 0 & {\cellcolor[HTML]{89B1D4}} \color[HTML]{000000} 29 \\
\bfseries Both Happening (Diff. Part) & {\cellcolor[HTML]{EEE9F3}} \color[HTML]{000000} 2 & {\cellcolor[HTML]{D4D4E8}} \color[HTML]{000000} 3 & {\cellcolor[HTML]{F0EAF4}} \color[HTML]{000000} 2 & {\cellcolor[HTML]{1379B5}} \color[HTML]{F1F1F1} 5 & {\cellcolor[HTML]{FFF7FB}} \color[HTML]{000000} 0 & {\cellcolor[HTML]{ECE7F2}} \color[HTML]{000000} 1 & {\cellcolor[HTML]{E3E0EE}} \color[HTML]{000000} 2 & {\cellcolor[HTML]{023858}} \color[HTML]{F1F1F1} 2 & {\cellcolor[HTML]{DDDBEC}} \color[HTML]{000000} 17 \\
\bfseries Gold Error (Correct) & {\cellcolor[HTML]{D7D6E9}} \color[HTML]{000000} 3 & {\cellcolor[HTML]{FFF7FB}} \color[HTML]{000000} 0 & {\cellcolor[HTML]{DBDAEB}} \color[HTML]{000000} 3 & {\cellcolor[HTML]{FFF7FB}} \color[HTML]{000000} 0 & {\cellcolor[HTML]{023858}} \color[HTML]{F1F1F1} 18 & {\cellcolor[HTML]{73A9CF}} \color[HTML]{F1F1F1} 4 & {\cellcolor[HTML]{2685BB}} \color[HTML]{F1F1F1} 5 & {\cellcolor[HTML]{FFF7FB}} \color[HTML]{000000} 0 & {\cellcolor[HTML]{63A2CB}} \color[HTML]{F1F1F1} 33 \\
\bfseries Gold Error (Incorrect) & {\cellcolor[HTML]{FFF7FB}} \color[HTML]{000000} 1 & {\cellcolor[HTML]{FFF7FB}} \color[HTML]{000000} 0 & {\cellcolor[HTML]{F0EAF4}} \color[HTML]{000000} 2 & {\cellcolor[HTML]{E8E4F0}} \color[HTML]{000000} 1 & {\cellcolor[HTML]{EEE9F3}} \color[HTML]{000000} 2 & {\cellcolor[HTML]{FFF7FB}} \color[HTML]{000000} 0 & {\cellcolor[HTML]{FFF7FB}} \color[HTML]{000000} 1 & {\cellcolor[HTML]{73A9CF}} \color[HTML]{F1F1F1} 1 & {\cellcolor[HTML]{FFF7FB}} \color[HTML]{000000} 8 \\
\bfseries True for Previous & {\cellcolor[HTML]{D7D6E9}} \color[HTML]{000000} 3 & {\cellcolor[HTML]{BDC8E1}} \color[HTML]{000000} 4 & {\cellcolor[HTML]{FFF7FB}} \color[HTML]{000000} 1 & {\cellcolor[HTML]{C4CBE3}} \color[HTML]{000000} 2 & {\cellcolor[HTML]{E3E0EE}} \color[HTML]{000000} 3 & {\cellcolor[HTML]{D0D1E6}} \color[HTML]{000000} 2 & {\cellcolor[HTML]{E3E0EE}} \color[HTML]{000000} 2 & {\cellcolor[HTML]{73A9CF}} \color[HTML]{F1F1F1} 1 & {\cellcolor[HTML]{D9D8EA}} \color[HTML]{000000} 18 \\
\bfseries Predicted Other & {\cellcolor[HTML]{023858}} \color[HTML]{F1F1F1} 10 & {\cellcolor[HTML]{023858}} \color[HTML]{F1F1F1} 13 & {\cellcolor[HTML]{023858}} \color[HTML]{F1F1F1} 11 & {\cellcolor[HTML]{1379B5}} \color[HTML]{F1F1F1} 5 & {\cellcolor[HTML]{FFF7FB}} \color[HTML]{000000} 0 & {\cellcolor[HTML]{023858}} \color[HTML]{F1F1F1} 8 & {\cellcolor[HTML]{023858}} \color[HTML]{F1F1F1} 7 & {\cellcolor[HTML]{73A9CF}} \color[HTML]{F1F1F1} 1 & {\cellcolor[HTML]{023858}} \color[HTML]{F1F1F1} 55 \\
\bfseries No Idea & {\cellcolor[HTML]{FFF7FB}} \color[HTML]{000000} 1 & {\cellcolor[HTML]{D4D4E8}} \color[HTML]{000000} 3 & {\cellcolor[HTML]{DBDAEB}} \color[HTML]{000000} 3 & {\cellcolor[HTML]{1379B5}} \color[HTML]{F1F1F1} 5 & {\cellcolor[HTML]{EEE9F3}} \color[HTML]{000000} 2 & {\cellcolor[HTML]{A5BDDB}} \color[HTML]{000000} 3 & {\cellcolor[HTML]{B4C4DF}} \color[HTML]{000000} 3 & {\cellcolor[HTML]{FFF7FB}} \color[HTML]{000000} 0 & {\cellcolor[HTML]{CED0E6}} \color[HTML]{000000} 20 \\
\midrule
\bfseries Total & {\cellcolor[HTML]{FFFFFF}} \color[HTML]{000000} 25 & {\cellcolor[HTML]{FFFFFF}} \color[HTML]{000000} 25 & {\cellcolor[HTML]{FFFFFF}} \color[HTML]{000000} 25 & {\cellcolor[HTML]{FFFFFF}} \color[HTML]{000000} 25 & {\cellcolor[HTML]{FFFFFF}} \color[HTML]{000000} 25 & {\cellcolor[HTML]{FFFFFF}} \color[HTML]{000000} 25 & {\cellcolor[HTML]{FFFFFF}} \color[HTML]{000000} 25 & {\cellcolor[HTML]{FFFFFF}} \color[HTML]{000000} 5 & {\cellcolor[HTML]{FFFFFF}} \color[HTML]{000000} 180 \\
\bottomrule
\end{tabular}
}
\caption{\FOS error counts by gold label and error category.}
\label{tab:error-counts}
\end{table*}

\subsection{Results} \label{sec:da-results}
The evaluation results for each of these model variants are summarized in Table \ref{tab:da-results}. While models incorporating dialog act data generally outperform the baseline, they do not improve upon the \FOS and, in fact, hamper performance across the board for these high-level measures. At first glance, this is a puzzling negative result. However, inspecting the performance on minority classes (see Table \ref{tab:da-results-min}) we find several instances where SWDA, the more granular of the two, improves recall while MRDA, which uses a coarser tag set, improves precision.

To make sense of these distinctions we employ a set of non-parametric statistical tests. This minimizes the assumptions made about our data, as no distribution is required, while still allowing us to assign rankings to the variables under question. Holding the training regimen as a treatment variable we utilize the Friedman rank sum test to determine if this contributes significantly to predicting model F1, precision, and recall by considering the null hypothesis that all variations contribute equally. This, along with Kendell's W, identifies where the models are performing differently from each other and the effect size of that difference, respectively. Significant differences on the minority classes can be seen in Table \ref{tab:da-results-min}. According to Cohen's interpretation guidelines for Kendall's W (0.1: small effect, 0.3: moderate effect, > 0.5: large effect) we find moderate to large effects in all metrics with significant differences. This confirms our suspicion that the dialog acts were indeed helping for minority labels. % We then employ the Neymenyi test to determine where exactly these identified differences occur.

This raises the question of what exactly the dialog act data was providing to the model and why did it succeed in some cases but fail to improve overall performance. In the subsequent sections, we perform an extensive error analysis which carefully examines our data and results to better disentangle the possible causes.

\section{Error Analysis and Discussion} \label{sec:error-analysis}

% We look at three basic questions
% What is the Fos system doing
% Does the DA data actually have any signal
% Are the DA models using that signal
% To which the short answers are mixing up face/probability matching, yes, and yes.

Our error analysis seeks to answer three questions. (1) What is the \FOS doing? (2) Is there signal in the dialog act data being incorporated and if there is, (3) are our systems that are trained with dialog act information using that signal?
We start by making some high level observations (\S\ref{sec:basic-observations}), then perform a manual analysis of the output produced by \FOS (\S\ref{sec:error-manual}), and finally discuss the contribution dialog act tags were making by comparing output between the systems (\S\ref{sec:da-contribution}).

\subsection{Basic Observations} \label{sec:basic-observations}
% \owenshort{I am not wedded to this table or this subsection, just plunking down for now}
% We summarize the distribution of labels and classification results in Table~\ref{tab:fos-results} and, as expected, performance is highly correlated with frequency of a tag. 
As noted in Section~\ref{sec:fos-results} and shown in Table~\ref{tab:fos-results}, \FOS performance is highly correlated ($r = 0.77$) with the frequency of a tag.
For the four tags that occur more than 1,000 times, we obtain 73\% F-measure or more, and for three tags that occur between 250 and 350 times, we obtain F-measures below 56\%.
% \owenshort{We should probably remove spos- from consideration earlier, maybe in the \FOS results subsection}
Using this information we divide the labels into three classes: The majority classes (\Other, \HPosR, \SPosR, \HNegT), the minority classes (\HPosT, \HNegR, \SNegR), and the extremely rare (\SPosT). As discussed in Section \ref{sec:approach-eval}, we disregard \SPosT in our analysis as its infrequency causes highly volatile results.

The confusion matrix for \FOS (see Figure \ref{fig:basic-confusion}) shows that, like in previous work \citep{dutt-etal-2020-keeping}, the majority of misclassifications occur when no face act is predicted to be present (the \Other column) though one is indeed there.  As shown in Table~\ref{tab:fos-results}, \Other is the most frequent label in our corpus.  The next largest source of error is \HPosR, which is the second most common label in the data set.  We conclude that the model is, to some extent, probability matching.  Furthermore, the confusion matrix in Figure \ref{fig:basic-confusion} reveals that the system may struggle with distinguishing positive and negative face, with \HNegR being mistaken for \HPosR 18\% of the time, and \SNegR being mistaken for \SPosR 11\% of the time.

% \SPosT is so rare that the results are somewhat random and we disregard it.

% \begin{figure*}[!t]
%     \centering
%     \begin{subfigure}[c]{0.60\textwidth}
%     \includegraphics[width=\textwidth]{images/errror-counts.png}
%     \end{subfigure}
%     \begin{subfigure}[c]{0.39\textwidth}
%     \includegraphics[width=\textwidth]{images/error-counts-pie.png}
%     \end{subfigure}
%     \caption{
%         \FOS error category percentages (right) and counts per category broken down by reference label (left).
%     }
%     \label{fig:error-counts}
% \end{figure*}

\begin{table*}[!htb]
\centering
\resizebox{\linewidth}{!}{
\begin{tabular}{lrrrrrrrr}
\toprule
 & \bfseries \SPosR & \bfseries \HPosR & \bfseries \SPosT & \bfseries \HPosT & \bfseries \SNegR & \bfseries \HNegR & \bfseries \HNegT & \bfseries \Other \\
\midrule
\bfseries BackChannel & {\cellcolor[HTML]{C7D7F0}} \color[HTML]{000000} -0.02 & {\cellcolor[HTML]{CEDAEB}} \color[HTML]{000000} 0.01 & {\cellcolor[HTML]{CCD9ED}} \color[HTML]{000000} -0.00 & {\cellcolor[HTML]{CAD8EF}} \color[HTML]{000000} -0.01 & {\cellcolor[HTML]{CBD8EE}} \color[HTML]{000000} -0.01 & {\cellcolor[HTML]{CAD8EF}} \color[HTML]{000000} -0.01 & {\cellcolor[HTML]{C9D7F0}} \color[HTML]{000000} -0.01 & {\cellcolor[HTML]{D3DBE7}} \color[HTML]{000000} 0.02 \\
\bfseries Disruption & {\cellcolor[HTML]{D5DBE5}} \color[HTML]{000000} 0.03 & {\cellcolor[HTML]{CEDAEB}} \color[HTML]{000000} 0.01 & {\cellcolor[HTML]{CBD8EE}} \color[HTML]{000000} -0.00 & {\cellcolor[HTML]{CBD8EE}} \color[HTML]{000000} -0.00 & {\cellcolor[HTML]{D4DBE6}} \color[HTML]{000000} 0.02 & {\cellcolor[HTML]{CBD8EE}} \color[HTML]{000000} -0.01 & {\cellcolor[HTML]{BCD2F7}} \color[HTML]{000000} -0.04 & {\cellcolor[HTML]{CBD8EE}} \color[HTML]{000000} -0.00 \\
\bfseries FloorGrabber & {\cellcolor[HTML]{C7D7F0}} \color[HTML]{000000} -0.01 & {\cellcolor[HTML]{C6D6F1}} \color[HTML]{000000} -0.02 & {\cellcolor[HTML]{CCD9ED}} \color[HTML]{000000} -0.00 & {\cellcolor[HTML]{CBD8EE}} \color[HTML]{000000} -0.01 & {\cellcolor[HTML]{CBD8EE}} \color[HTML]{000000} -0.01 & {\cellcolor[HTML]{CBD8EE}} \color[HTML]{000000} -0.01 & {\cellcolor[HTML]{C9D7F0}} \color[HTML]{000000} -0.01 & {\cellcolor[HTML]{D9DCE1}} \color[HTML]{000000} 0.04 \\
\bfseries Question & {\cellcolor[HTML]{84A7FC}} \color[HTML]{F1F1F1} -0.19 & {\cellcolor[HTML]{6E90F2}} \color[HTML]{F1F1F1} -0.24 & {\cellcolor[HTML]{C6D6F1}} \color[HTML]{000000} -0.02 & {\cellcolor[HTML]{D4DBE6}} \color[HTML]{000000} 0.02 & {\cellcolor[HTML]{B3CDFB}} \color[HTML]{000000} -0.07 & {\cellcolor[HTML]{B9D0F9}} \color[HTML]{000000} -0.05 & {\cellcolor[HTML]{B40426}} \color[HTML]{F1F1F1} 0.49 & {\cellcolor[HTML]{E7D7CE}} \color[HTML]{000000} 0.09 \\
\bfseries Statement & {\cellcolor[HTML]{F2C9B4}} \color[HTML]{000000} 0.14 & {\cellcolor[HTML]{F7B599}} \color[HTML]{000000} 0.20 & {\cellcolor[HTML]{D2DBE8}} \color[HTML]{000000} 0.02 & {\cellcolor[HTML]{C6D6F1}} \color[HTML]{000000} -0.02 & {\cellcolor[HTML]{DADCE0}} \color[HTML]{000000} 0.04 & {\cellcolor[HTML]{DCDDDD}} \color[HTML]{000000} 0.05 & {\cellcolor[HTML]{3B4CC0}} \color[HTML]{F1F1F1} -0.38 & {\cellcolor[HTML]{B1CBFC}} \color[HTML]{000000} -0.08 \\
\midrule
\bfseries Acknowledge (Backchannel) & {\cellcolor[HTML]{C0D4F5}} \color[HTML]{000000} -0.03 & {\cellcolor[HTML]{CCD9ED}} \color[HTML]{000000} 0.00 & {\cellcolor[HTML]{CBD8EE}} \color[HTML]{000000} -0.00 & {\cellcolor[HTML]{C7D7F0}} \color[HTML]{000000} -0.02 & {\cellcolor[HTML]{C7D7F0}} \color[HTML]{000000} -0.01 & {\cellcolor[HTML]{CBD8EE}} \color[HTML]{000000} -0.00 & {\cellcolor[HTML]{C1D4F4}} \color[HTML]{000000} -0.03 & {\cellcolor[HTML]{DDDCDC}} \color[HTML]{000000} 0.05 \\
\bfseries Action-directive & {\cellcolor[HTML]{CAD8EF}} \color[HTML]{000000} -0.01 & {\cellcolor[HTML]{C7D7F0}} \color[HTML]{000000} -0.01 & {\cellcolor[HTML]{CCD9ED}} \color[HTML]{000000} -0.00 & {\cellcolor[HTML]{CDD9EC}} \color[HTML]{000000} 0.00 & {\cellcolor[HTML]{CAD8EF}} \color[HTML]{000000} -0.01 & {\cellcolor[HTML]{CAD8EF}} \color[HTML]{000000} -0.01 & {\cellcolor[HTML]{D9DCE1}} \color[HTML]{000000} 0.04 & {\cellcolor[HTML]{CBD8EE}} \color[HTML]{000000} -0.00 \\
\bfseries Affirmative non-yes answers & {\cellcolor[HTML]{CEDAEB}} \color[HTML]{000000} 0.00 & {\cellcolor[HTML]{D7DCE3}} \color[HTML]{000000} 0.03 & {\cellcolor[HTML]{CCD9ED}} \color[HTML]{000000} -0.00 & {\cellcolor[HTML]{CAD8EF}} \color[HTML]{000000} -0.01 & {\cellcolor[HTML]{CAD8EF}} \color[HTML]{000000} -0.01 & {\cellcolor[HTML]{CAD8EF}} \color[HTML]{000000} -0.01 & {\cellcolor[HTML]{C7D7F0}} \color[HTML]{000000} -0.01 & {\cellcolor[HTML]{C6D6F1}} \color[HTML]{000000} -0.02 \\
\bfseries Agree/Accept & {\cellcolor[HTML]{C4D5F3}} \color[HTML]{000000} -0.02 & {\cellcolor[HTML]{F3C7B1}} \color[HTML]{000000} 0.15 & {\cellcolor[HTML]{CBD8EE}} \color[HTML]{000000} -0.00 & {\cellcolor[HTML]{C5D6F2}} \color[HTML]{000000} -0.02 & {\cellcolor[HTML]{C5D6F2}} \color[HTML]{000000} -0.02 & {\cellcolor[HTML]{C5D6F2}} \color[HTML]{000000} -0.02 & {\cellcolor[HTML]{BCD2F7}} \color[HTML]{000000} -0.04 & {\cellcolor[HTML]{B2CCFB}} \color[HTML]{000000} -0.07 \\
\bfseries Appreciation & {\cellcolor[HTML]{9FBFFF}} \color[HTML]{000000} -0.12 & {\cellcolor[HTML]{F18F71}} \color[HTML]{F1F1F1} 0.29 & {\cellcolor[HTML]{CFDAEA}} \color[HTML]{000000} 0.01 & {\cellcolor[HTML]{BAD0F8}} \color[HTML]{000000} -0.05 & {\cellcolor[HTML]{BBD1F8}} \color[HTML]{000000} -0.05 & {\cellcolor[HTML]{BBD1F8}} \color[HTML]{000000} -0.05 & {\cellcolor[HTML]{A6C4FE}} \color[HTML]{000000} -0.10 & {\cellcolor[HTML]{B6CEFA}} \color[HTML]{000000} -0.06 \\
\bfseries Backchannel in question form & {\cellcolor[HTML]{C7D7F0}} \color[HTML]{000000} -0.01 & {\cellcolor[HTML]{C7D7F0}} \color[HTML]{000000} -0.01 & {\cellcolor[HTML]{CCD9ED}} \color[HTML]{000000} -0.00 & {\cellcolor[HTML]{CAD8EF}} \color[HTML]{000000} -0.01 & {\cellcolor[HTML]{CBD8EE}} \color[HTML]{000000} -0.01 & {\cellcolor[HTML]{CBD8EE}} \color[HTML]{000000} -0.01 & {\cellcolor[HTML]{CBD8EE}} \color[HTML]{000000} -0.00 & {\cellcolor[HTML]{D7DCE3}} \color[HTML]{000000} 0.03 \\
\bfseries Conventional-closing & {\cellcolor[HTML]{ABC8FD}} \color[HTML]{000000} -0.09 & {\cellcolor[HTML]{DDDCDC}} \color[HTML]{000000} 0.06 & {\cellcolor[HTML]{CEDAEB}} \color[HTML]{000000} 0.01 & {\cellcolor[HTML]{BFD3F6}} \color[HTML]{000000} -0.04 & {\cellcolor[HTML]{C3D5F4}} \color[HTML]{000000} -0.03 & {\cellcolor[HTML]{C0D4F5}} \color[HTML]{000000} -0.03 & {\cellcolor[HTML]{B2CCFB}} \color[HTML]{000000} -0.07 & {\cellcolor[HTML]{E9D5CB}} \color[HTML]{000000} 0.09 \\
\bfseries Conventional-opening & {\cellcolor[HTML]{C9D7F0}} \color[HTML]{000000} -0.01 & {\cellcolor[HTML]{C6D6F1}} \color[HTML]{000000} -0.02 & {\cellcolor[HTML]{CCD9ED}} \color[HTML]{000000} -0.00 & {\cellcolor[HTML]{CBD8EE}} \color[HTML]{000000} -0.01 & {\cellcolor[HTML]{CBD8EE}} \color[HTML]{000000} -0.00 & {\cellcolor[HTML]{CBD8EE}} \color[HTML]{000000} -0.00 & {\cellcolor[HTML]{C9D7F0}} \color[HTML]{000000} -0.01 & {\cellcolor[HTML]{D8DCE2}} \color[HTML]{000000} 0.04 \\
\bfseries Hedge & {\cellcolor[HTML]{C9D7F0}} \color[HTML]{000000} -0.01 & {\cellcolor[HTML]{CBD8EE}} \color[HTML]{000000} -0.01 & {\cellcolor[HTML]{CCD9ED}} \color[HTML]{000000} -0.00 & {\cellcolor[HTML]{CBD8EE}} \color[HTML]{000000} -0.00 & {\cellcolor[HTML]{D4DBE6}} \color[HTML]{000000} 0.02 & {\cellcolor[HTML]{CBD8EE}} \color[HTML]{000000} -0.00 & {\cellcolor[HTML]{CAD8EF}} \color[HTML]{000000} -0.01 & {\cellcolor[HTML]{D1DAE9}} \color[HTML]{000000} 0.01 \\
\bfseries Hold before answer/agreement & {\cellcolor[HTML]{C5D6F2}} \color[HTML]{000000} -0.02 & {\cellcolor[HTML]{CAD8EF}} \color[HTML]{000000} -0.01 & {\cellcolor[HTML]{CCD9ED}} \color[HTML]{000000} -0.00 & {\cellcolor[HTML]{CAD8EF}} \color[HTML]{000000} -0.01 & {\cellcolor[HTML]{CAD8EF}} \color[HTML]{000000} -0.01 & {\cellcolor[HTML]{CAD8EF}} \color[HTML]{000000} -0.01 & {\cellcolor[HTML]{C6D6F1}} \color[HTML]{000000} -0.02 & {\cellcolor[HTML]{DADCE0}} \color[HTML]{000000} 0.04 \\
\bfseries No answers & {\cellcolor[HTML]{C6D6F1}} \color[HTML]{000000} -0.02 & {\cellcolor[HTML]{CCD9ED}} \color[HTML]{000000} 0.00 & {\cellcolor[HTML]{CCD9ED}} \color[HTML]{000000} -0.00 & {\cellcolor[HTML]{D5DBE5}} \color[HTML]{000000} 0.03 & {\cellcolor[HTML]{CEDAEB}} \color[HTML]{000000} 0.01 & {\cellcolor[HTML]{CAD8EF}} \color[HTML]{000000} -0.01 & {\cellcolor[HTML]{C7D7F0}} \color[HTML]{000000} -0.02 & {\cellcolor[HTML]{D1DAE9}} \color[HTML]{000000} 0.01 \\
\bfseries Non-verbal & {\cellcolor[HTML]{B7CFF9}} \color[HTML]{000000} -0.06 & {\cellcolor[HTML]{CEDAEB}} \color[HTML]{000000} 0.00 & {\cellcolor[HTML]{D7DCE3}} \color[HTML]{000000} 0.03 & {\cellcolor[HTML]{C6D6F1}} \color[HTML]{000000} -0.02 & {\cellcolor[HTML]{D3DBE7}} \color[HTML]{000000} 0.02 & {\cellcolor[HTML]{CDD9EC}} \color[HTML]{000000} 0.00 & {\cellcolor[HTML]{C3D5F4}} \color[HTML]{000000} -0.03 & {\cellcolor[HTML]{DDDCDC}} \color[HTML]{000000} 0.05 \\
\bfseries Other & {\cellcolor[HTML]{C0D4F5}} \color[HTML]{000000} -0.03 & {\cellcolor[HTML]{BBD1F8}} \color[HTML]{000000} -0.05 & {\cellcolor[HTML]{CBD8EE}} \color[HTML]{000000} -0.00 & {\cellcolor[HTML]{C7D7F0}} \color[HTML]{000000} -0.01 & {\cellcolor[HTML]{C7D7F0}} \color[HTML]{000000} -0.01 & {\cellcolor[HTML]{C7D7F0}} \color[HTML]{000000} -0.01 & {\cellcolor[HTML]{C3D5F4}} \color[HTML]{000000} -0.03 & {\cellcolor[HTML]{EAD4C8}} \color[HTML]{000000} 0.10 \\
\bfseries Other answers & {\cellcolor[HTML]{CBD8EE}} \color[HTML]{000000} -0.00 & {\cellcolor[HTML]{CBD8EE}} \color[HTML]{000000} -0.01 & {\cellcolor[HTML]{CCD9ED}} \color[HTML]{000000} -0.00 & {\cellcolor[HTML]{CCD9ED}} \color[HTML]{000000} -0.00 & {\cellcolor[HTML]{CCD9ED}} \color[HTML]{000000} -0.00 & {\cellcolor[HTML]{CCD9ED}} \color[HTML]{000000} -0.00 & {\cellcolor[HTML]{CBD8EE}} \color[HTML]{000000} -0.00 & {\cellcolor[HTML]{D1DAE9}} \color[HTML]{000000} 0.01 \\
\bfseries Response Acknowledgement & {\cellcolor[HTML]{C7D7F0}} \color[HTML]{000000} -0.01 & {\cellcolor[HTML]{CCD9ED}} \color[HTML]{000000} -0.00 & {\cellcolor[HTML]{CCD9ED}} \color[HTML]{000000} -0.00 & {\cellcolor[HTML]{CBD8EE}} \color[HTML]{000000} -0.01 & {\cellcolor[HTML]{CBD8EE}} \color[HTML]{000000} -0.01 & {\cellcolor[HTML]{CBD8EE}} \color[HTML]{000000} -0.01 & {\cellcolor[HTML]{C9D7F0}} \color[HTML]{000000} -0.01 & {\cellcolor[HTML]{D4DBE6}} \color[HTML]{000000} 0.02 \\
\bfseries Segment & {\cellcolor[HTML]{E3D9D3}} \color[HTML]{000000} 0.07 & {\cellcolor[HTML]{CEDAEB}} \color[HTML]{000000} 0.01 & {\cellcolor[HTML]{CAD8EF}} \color[HTML]{000000} -0.01 & {\cellcolor[HTML]{CCD9ED}} \color[HTML]{000000} -0.00 & {\cellcolor[HTML]{D3DBE7}} \color[HTML]{000000} 0.02 & {\cellcolor[HTML]{D2DBE8}} \color[HTML]{000000} 0.02 & {\cellcolor[HTML]{BBD1F8}} \color[HTML]{000000} -0.05 & {\cellcolor[HTML]{BFD3F6}} \color[HTML]{000000} -0.04 \\
\bfseries Statement-non-opinion & {\cellcolor[HTML]{F08A6C}} \color[HTML]{F1F1F1} 0.30 & {\cellcolor[HTML]{A9C6FD}} \color[HTML]{000000} -0.10 & {\cellcolor[HTML]{D1DAE9}} \color[HTML]{000000} 0.01 & {\cellcolor[HTML]{D1DAE9}} \color[HTML]{000000} 0.01 & {\cellcolor[HTML]{EFCFBF}} \color[HTML]{000000} 0.12 & {\cellcolor[HTML]{E6D7CF}} \color[HTML]{000000} 0.08 & {\cellcolor[HTML]{799CF8}} \color[HTML]{F1F1F1} -0.21 & {\cellcolor[HTML]{B1CBFC}} \color[HTML]{000000} -0.07 \\
\bfseries Statement-opinion & {\cellcolor[HTML]{C1D4F4}} \color[HTML]{000000} -0.03 & {\cellcolor[HTML]{F1CDBA}} \color[HTML]{000000} 0.13 & {\cellcolor[HTML]{C9D7F0}} \color[HTML]{000000} -0.01 & {\cellcolor[HTML]{DBDCDE}} \color[HTML]{000000} 0.05 & {\cellcolor[HTML]{C0D4F5}} \color[HTML]{000000} -0.03 & {\cellcolor[HTML]{D3DBE7}} \color[HTML]{000000} 0.02 & {\cellcolor[HTML]{A9C6FD}} \color[HTML]{000000} -0.10 & {\cellcolor[HTML]{BBD1F8}} \color[HTML]{000000} -0.05 \\
\bfseries Wh-Question & {\cellcolor[HTML]{96B7FF}} \color[HTML]{000000} -0.14 & {\cellcolor[HTML]{88ABFD}} \color[HTML]{000000} -0.18 & {\cellcolor[HTML]{C9D7F0}} \color[HTML]{000000} -0.01 & {\cellcolor[HTML]{D4DBE6}} \color[HTML]{000000} 0.02 & {\cellcolor[HTML]{BAD0F8}} \color[HTML]{000000} -0.05 & {\cellcolor[HTML]{BBD1F8}} \color[HTML]{000000} -0.05 & {\cellcolor[HTML]{E36C55}} \color[HTML]{F1F1F1} 0.35 & {\cellcolor[HTML]{E3D9D3}} \color[HTML]{000000} 0.07 \\
\bfseries Yes answers & {\cellcolor[HTML]{CEDAEB}} \color[HTML]{000000} 0.01 & {\cellcolor[HTML]{D1DAE9}} \color[HTML]{000000} 0.01 & {\cellcolor[HTML]{CCD9ED}} \color[HTML]{000000} -0.00 & {\cellcolor[HTML]{C7D7F0}} \color[HTML]{000000} -0.01 & {\cellcolor[HTML]{CBD8EE}} \color[HTML]{000000} -0.00 & {\cellcolor[HTML]{CBD8EE}} \color[HTML]{000000} -0.01 & {\cellcolor[HTML]{C4D5F3}} \color[HTML]{000000} -0.03 & {\cellcolor[HTML]{CFDAEA}} \color[HTML]{000000} 0.01 \\
\bfseries Yes-No-Question & {\cellcolor[HTML]{A1C0FF}} \color[HTML]{000000} -0.12 & {\cellcolor[HTML]{90B2FE}} \color[HTML]{000000} -0.16 & {\cellcolor[HTML]{C9D7F0}} \color[HTML]{000000} -0.01 & {\cellcolor[HTML]{CFDAEA}} \color[HTML]{000000} 0.01 & {\cellcolor[HTML]{BCD2F7}} \color[HTML]{000000} -0.05 & {\cellcolor[HTML]{C3D5F4}} \color[HTML]{000000} -0.03 & {\cellcolor[HTML]{EE8669}} \color[HTML]{F1F1F1} 0.31 & {\cellcolor[HTML]{DFDBD9}} \color[HTML]{000000} 0.06 \\
\bottomrule
\end{tabular}
}
\caption{The correlation coefficients for the dialog act labels produced by \citet{he-etal-2021-speaker-turn}'s system for MRDA (top) and SWDA (bottom). These were incorporated into the training data for \TAM and \TAS (\S\ref{sec:methodology-da}).}
\label{tab:da-corrs}
\end{table*}

\subsection{Error Analysis on the \FOS} \label{sec:error-manual}

We perform manual analysis of the output produced by our best performing system (\FOS) by collecting 25 misclassified examples for each label with five randomly selected from each test fold for a total of up to 200 output-prediction pairs. As every test fold does not necessarily contain five incorrect predictions for \SPosT, this produced 180 examples for study.

We sort these examples into the following seven error categories and consult the annotation guidelines and criteria to make the appropriate determinations.
{\fontsize{10.5}{12}\selectfont
\begin{enumerate}
    \item \tbf{Both Happening (Same Part):} The predicted face act is also happening for that utterance in the same span of text (two face acts at once).
    \item \tbf{Both Happening (Diff. Part):} The predicted face act is also happening for that utterance, but in a different span of the text.
    \item \tbf{Gold Error (Correct):} The reference face act label is incorrect and the predicted face act label is correct.
    \item \tbf{Gold Error (Incorrect):} The reference face act label is incorrect and the predicted face act label is also incorrect.
    \item \tbf{True for Previous:} The predicted face act occurs for a previous utterance in the context window.
    \item \tbf{Predicted Other:} None of the previous error categories apply and the system predicted \Other.
    \item \tbf{No Idea:} None of the previous error categories apply and we could not determine a specific reason for this (errorful) prediction.
\end{enumerate}
}

The results of this analysis are summarized in Table~\ref{tab:error-counts}. Overall, we find a gold error in 22.7\% of the examples examined with 18.3\% absolute of these errors being identified as instances where the system output was correct. In another 25.5\% of cases we identified criteria in the annotation guidelines for both the gold label {\em and} the predicted label (meaning that both labels are correct), with the ``same'' and ``different'' part categories occurring with roughly equal frequency. This is in stark contrast to the reported 2\% frequency of multi-label annotations observed by \citet{dutt-etal-2020-keeping}. 10\% of the time the system produced a prediction which is correct for a previous utterance in the context.
This points to a systematic issue with this method of integrating the context and suggests performance could be improved by helping the model more accurately identify the exact utterance under question. % XXX: Added
Among the cases in which no discernible pattern was identified, 30.6\% absolute involved the system predicting no face act to be present (i.e., \Other), and 11.2\% absolute involved another prediction.  Thus, in summary, we have a gold error rate of 22.7\%, and we find that in 43.8\% of errors the predicted face act is actually correct (possibly among others).

We also break these counts down by gold label and error category in Table~\ref{tab:error-counts}. 
% Note that the blue, orange, and green parts of the table refer to predictions that are actually correct.\footnote{We apologize to reviewers who have trouble identifying these colors or who are using gray-scale technology.}  
% === NEW
Note that the first three rows of the table refer to predictions that are actually correct.
% === NEW
The analysis reveals that the system's prediction is in fact correct for a majority of the cases with gold labels of \Other, and about half the time for cases in which the gold label is \HPosT, \SNegR, or \SPosR.  The worst performance is on \HNegT.  
% Furthermore, the predicted label is correct in almost all of these cases (i.e., we have a gold error). In half of the examples with gold label \HPosT we identify criteria for the predicted face act as well.  
%Across all face acts the system produces labels which are correct for a previous utterance in the context with roughly the same frequency. 

% Error analysis of the best system. (sweep-xval)
% - Look at (up to) 25 incorrect examples (5 randomly selected from each test fold) for each class (total of 200).
% - Classify errors by categories:
%   1) Data issues
%     - Gold Error (using their criteria)
%     - True according to annotation but not the one selected from the data
%     - Possible to us (but not according to the annotation criteria)
%   2) Prediction errors
%     - Makes sense but not in this context (?)
%     - Nonsense
%     - Any other trend that jumps out.  <-- 

% Automatically on all data:
%     - Analysis by components (H/S, Pos/Neg, +/-)
%     - Confusion matrix

\subsection{Contribution of Dialog Acts} \label{sec:da-contribution}
% Start with correlation table, which shows that the basic idea is correct

% Hypothesize that the FTA tagger learns some relevant DA distinctions so that they do not help (question/statement for xxx)

% Task: find all cases where gold OR xval prediction OR xval-with-swda is hneg+ and where xval and xval-with-swda differ (columns: [sentence, gold-label, with-swda-label, without-swda-label, da-label])

% Find where non-xval systems perform better

% - look at where the da system gets right and other one doesn't.
% - find an example or two
% - hypothesize explanation

% Observations we should make:

% 1) hneg+ story: with models perform much better than all others\owenshort{I have some numbers -- will fill in text when there is more above I can refer to!}

% hneg+ is an outlier, and ONLY the 
% WIth swda helps with recall on hneg+

% 2) remaing MIN story:

% MTL swda helps with recall on ALL 4 minority classes EXCEPT hneg+.

\begin{table*}[!ht]
    \centering
    \scalebox{1}{
    \begin{tabular}{lcccccc}
        \toprule
        & \bfseries All  & \bfseries \HNegR & \bfseries FN to TP & \bfseries TP to FN & \bfseries FP to TN & \bfseries TN to FP \\
        \midrule
        \bfseries Statement & 80\% & 93\% & 97\% & 100\% & 78\% & 91\% \\ 
        \bfseries Question & 20\% & 7\% & 3\% & 0\% & 22\% & 9\% \\
        \midrule
        \bfseries Count & - & 305 & 64 & 20 & 50 & 101 \\
        \bottomrule
    \end{tabular}}
    \caption{Distribution of MRDA dialog act tags Statement (containing also Disruption) and Question for different cases relating to predicting \HNegR, FN refers to false negatives, TP to true positives, and so on.}
    \label{tab:hneg}
\end{table*}

We now turn to investigating the effects of integrating dialog acts into the model. We do so by first examining correlations found between face acts and dialog acts included in the text-augmented model data to determine, roughly, where signal might be. The results of this analysis are shown in Table \ref{tab:da-corrs} and qualitatively support our hypothesis that these concepts, face and intention, are intimately related. Inspecting the MRDA tags, there is a strong positive correlation ($r = 0.49$) between questions and \HNegT with a correspondingly negative correlation ($r = -0.38$) for statements. The trend continues across the various question categories for the SWDA tags though, critically, no correlation is observed for backchannels which take the form of a question. As impositions on negative face often take the form of requests or questions rather than statements, with the exception of commands (which are not frequent in this corpus), these correlations are expected.
% These correlations are expected as impositions on negative face often take the form of requests or questions rather than statements, with the exception of commands which are not expected to be frequent in this corpus.
Furthermore, the trend is nicely reversed for \HPosR, which sees positive correlation ($r = 0.20$) with statements and negative ($r = -0.24$) with questions. The SWDA tags reveal this largely comes from the appreciation ($r = 0.29$) and agreement ($r = 0.15$) labels which are stereotypical examples of the \HPosR face act. Because the corpus contains a large number of utterances in which participants signal virtue (e.g. \emph{I often make large donations}), the correlation between \SPosR and non-opinion statements ($r = 0.30$) makes sense.

This all indicates that the dialog act data included in the text-augmented models, and learned from the same data set used in training the multi-task models, does contain information with signal. One might conclude that (1) dialog acts were not providing very much orthogonal information in training (i.e. \FOS already learned to distinguish these) or, if it was, perhaps (2) these methods of integration were not effective for this task.

% \begin{table*}[!ht]
%     \centering
%     \scalebox{1}{
%     \begin{tabular}{|l|l|l|l|l|l|l|}
%     \hline
%         & \bfseries All  & \bfseries \HNegR & \bfseries FN to TP & \bfseries TP to FN & \bfseries FP to TN & \bfseries TN to FP \\ \hline
%         \bfseries Statement & 80\% & 93\% & 97\% & 100\% & 78\% & 91\% \\ \hline
%         \bfseries Question & 20\% & 7\% & 3\% & 0\% & 22\% & 9\% \\ \hline
%         \bfseries Number &  & 305 & 64 & 20 & 50 & 101 \\ \hline
%     \end{tabular}}
%     \caption{Distribution of MRDA dialog act tags Statement (containing also Disruption) and Question for different cases relating to predicting \HNegR, FN refers to false negatives, TP to true positives, and so on.}
%     \label{tab:hneg}
% \end{table*}

To untangle these options, we carefully examine examples where the gold label or output from \FOS, \TAM, or \TAS contained \HNegR. This label indicates that an utterance aims to raise the hearer's negative face, meaning that the speaker is trying to provide or point out options for action to the hearer (rather than restrict them).  
We focus on this label for analysis because it is an outlier in that this is the only label for which the \FOS system did not obtain the best F1 result (see Table~\ref{tab:da-results-min}), meaning that dialog acts actually helped its performance.  We use the MRDA dialog acts to investigate this phenomenon, partly because the MRDA tag set is simpler than the SWDA tag set, and partly because the \TAM system obtained the best results of all five systems.  
%It did so by having a much higher recall than \FOS, while matching its precision.   
There are 305 instances of \HNegR in the corpus.  We investigate the distribution of dialog act tags.  In doing this, we collapse Statement and Disruption, as we could find no meaningful distinction in this written dialog corpus.  Since the other tags are exceedingly rare, we restrict our analysis to the distinction between Statement (to which we have added Disruption) and Question.  The results are in Table~\ref{tab:hneg}.

In the first two columns, we see that overall, 20\% of turns are questions, but for the 305 turns labeled \HNegR, only 7\% are, which is consistent with the semantics of the tag.  However, as we could see in Table~\ref{tab:da-corrs}, the correlation is not strong (5\% absolute for both dialog act tags). Despite the rather weak correlation, the presence of the tag increases the F1 measure from a low 44\% (the lowest of any tag) for the \FOS system to 51\% for the \TAM system, entirely by increasing recall from 41\% to 55\%.  To understand why, we can look at the cases in which the MRDA tags in the input change the prediction.  There are four cases: the predicted tag is \HNegR (Positive) or some other tag (Negative), and the prediction is correct (True) or not (False).  This gives us four possible shifts as we go from the \FOS system to the \TAM system which uses the dialog act tags.  The correct shifts from False Negative to True Positives have 97\% Statements, and the correct shift from False Positives to True Negatives have 22\% Questions -- as expected.  In terms of the incorrect shifts, the \TAM system does not change a True Positive to a False Negative frequently, and we disregard this case.  The 
\TAM system does change a True Negative to a False Positive very frequently (101 cases), but here the distribution of Statements and Questions is very close to that of \HNegR in general, so the dialog act label does not contribute to this error class.  In conclusion, we can see that simply adding a very simple dialog act distinction helps the classifier for a low-frequency tag in the way we expect, increasing TPs and decreasing FPs.

While it may seem odd that for the three face act tags for which the correlations are strongest (\SPosR, \HPosR, and \HNegT) we do not see an improvement from adding the tags, we note that these three face acts are also the most common in our corpus, and we assume that the \FOS system has enough data and can derive the face act tag from the lexical information on its own.

% "EE: Great to hear that you like our mission statement! (Statement)
% EE: Where would you like to focus your donation? (Question)
% EE: Toward children's health, education, or safety? (Question)"

% where the statistical testing conducted in Section \ref{sec:da-results} indicated the most significant and largest differences, particularly for the text-augmented models, suggesting there is something to be learned here.

\section{Conclusion and Future Work} \label{sec:conclusion}

We have presented a new study on the face act corpus of \citet{dutt-etal-2020-keeping} and use a generative approach to obtain state-of-the-art results. The model is then augmented to investigate the role of communicative intention in determining face acts. Through several methods of analysis we find evidence that there is a close relationship between dialog acts and face acts. Despite showing some improvement on minority labels (\S~\ref{sec:da-results}) and correlations between dialog acts and face acts (\S~\ref{sec:da-contribution}), our augmented models see an overall decline in performance when incorporating dialog acts. Our error analysis finds issues with the annotation consistency (\Other in particular) and, more importantly, points to methods of improving future annotation efforts.

We observe that some of the theoretical study that was done when developing dialog act representations is very applicable to face acts. Early work on speech act theory also limited utterances to a single label but this was, over time, identified as a serious flaw by several researchers \citep{cohen-levesque-1987-rational,hancher-1979-classification}. As a result, DAMSL \citep{core-and-allen-1997-coding}, now the most commonly adapted methodology for dialog act annotations, made supporting multiple labels a primary objective in its design. Future annotations of face acts should do the same and the frequency that multiple act utterances were found in our error analysis supports this recommendation.

% The takeaways of this paper can be divided by those relevant to linguistic theory and annotation, and those relevant to computational modeling and NLP. Our main results regarding the former are as follows. 
% From a linguistic standpoint our main results are as follow. (1) Finding empirical evidence to support the claim that there is a connection between dialog acts and face acts. (2) Identifying that some of the theoretical work that was done when developing dialog act representations is very applicable to face acts. Early work on speech act theory also limited utterances to a single label but this was, over time, identified as a serious flaw by several researchers \citep{cohen-levesque-1987-rational,hancher-1979-classification}. As a result, DAMSL \citep{core-and-allen-1997-coding}, now the most commonly adapted methodology for dialog act annotations, made supporting multiple labels a primary objective in its design.
% We feel that future annotations of face acts should do the same and the frequency that multiple act utterances were found in our error analysis supports this recommendation.

Face acts are an important part of language use, and while \citet{dutt-etal-2020-keeping} have made a major contribution, there has been little work with this corpus, and future work will require thinking hard about the data. We hope this paper will allow other researchers to use the corpus in constructive ways while being aware of the nature of the data in more detail. Despite their similarities from a computational perspective, it does not seem to be as simple as lifting the approaches used for dialog acts when modeling face acts.
% === NEW
In the future, we plan to develop an annotation which incorporates additional aspects of politeness theory and to label a corpus with in-house trained annotators. 
Once we have such a corpus, we predict that we will be able to exploit the double annotation of face acts and dialog acts in machine learning more effectively and obtain a much deeper understanding of how intention and modeling of the audience interact.
\section*{Ethics Statement}
The experiments for this work were performed using computational resources that are not, in general, freely available. In part due to these computational requirements, but also a result of minimal data, we were not able to evaluate the techniques on additional languages and acknowledge the limitations this places on extending our results to other cultures. We also note along similar lines that while \citet{brown-levinson-1987-politeness} claim their theory of politeness to be culturally universal, this claim has been contested -- most notably for eastern cultures \citep{al-duleimi-etal-2016-critical,purkarthofer-flubacher-2022-speaking}. As discussed in detail above, taking utterances to have a single face act or intent is a critically limiting assumption which lends some uncertainty to our conclusions.

% EMNLP 2023 requires all submissions to have a section titled ``Limitations'', for discussing the limitations of the paper as a complement to the discussion of strengths in the main text. This section should occur after the conclusion, but before the references. It will not count towards the page limit.  

% The discussion of limitations is mandatory. Papers without a limitation section will be desk-rejected without review.
% ARR-reviewed papers that did not include ``Limitations'' section in their prior submission, should submit a PDF with such a section together with their EMNLP 2023 submission.

% While we are open to different types of limitations, just mentioning that a set of results have been shown for English only probably does not reflect what we expect. 
% Mentioning that the method works mostly for languages with limited morphology, like English, is a much better alternative.
% In addition, limitations such as low scalability to long text, the requirement of large GPU resources, or other things that inspire crucial further investigation are welcome.

% \section*{Ethics Statement}
Despite a detailed analysis of the errors, we cannot verify the safety of this system in any user-oriented context and therefore do not recommend such uses without further study. While we do not produce any data sets directly from human annotations, we do use several which were, to the best of our knowledge, compiled ethically. As the primary object of study in this work is the relationship between face and intention, we do not anticipate broad risks to its application.

% Scientific work published at EMNLP 2023 must comply with the \href{https://www.aclweb.org/portal/content/acl-code-ethics}{ACL Ethics Policy}. We encourage all authors to include an explicit ethics statement on the broader impact of the work, or other ethical considerations after the conclusion but before the references. The ethics statement will not count toward the page limit (8 pages for long, 4 pages for short papers).

\section*{Acknowledgements}
% Add DARPA, IACS, AI Institute Acknowledgements here. Acknowledge Shyne. Acknowledge Reviewers.

This material is based upon work supported in part by the National Science Foundation (NSF) under No. 2125295 (NRT-HDR: Detecting and Addressing Bias in Data, Humans, and Institutions) as well as by funding from the Defense Advanced Research Projects Agency (DARPA) under the CCU (No. HR001120C0037, PR No. HR0011154158, No. HR001122C0034) program.  Any opinions, findings and conclusions or recommendations expressed in this material are those of the author(s) and do not necessarily reflect the views of the NSF or DARPA.

We thank both the Institute for Advanced Computational Science and the Institute for AI-Driven Discovery and Innovation at Stony Brook for access to the computing resources needed for this work. These resources were made possible by NSF grant No. 1531492 (SeaWulf HPC cluster maintained by Research Computing and Cyberinfrastructure) and NSF grant No. 1919752 (Major Research Infrastructure program), respectively.

We would also like to thank our anonymous reviewers for their perceptive comments, which improved this work, and Shyne Choi for her participation in several conversations which encouraged this study.

% no acks in submissions

% \begin{table*}[htb]
%     \centering
%     \resizebox{\textwidth}{!}{
%     \begin{tabular}{r | cccccccc | cc}
%     \toprule
%     Model &
%     \SPosR & \HPosR & \HPosT & \SNegR & \SNegT & \HNegR & \HNegT & \Other &
%     F1 & Macro F1 \\
%     \midrule
%     T5-Base & xx & xx & xx & xx & xx & xx & xx & xx & xx & xx \\
%     +MRDA & xx & xx & xx & xx & xx & xx & xx & xx & xx & xx \\
%     +SWDA & xx & xx & xx & xx & xx & xx & xx & xx & xx & xx \\
%     +GPT & xx & xx & xx & xx & xx & xx & xx & xx & xx & xx \\
%     \midrule
%     Flan-T5-Base & xx & xx & xx & xx & xx & xx & xx & xx & xx & xx \\
%     +MRDA & xx & xx & xx & xx & xx & xx & xx & xx & xx & xx \\
%     +SWDA & xx & xx & xx & xx & xx & xx & xx & xx & xx & xx \\
%     +GPT & xx & xx & xx & xx & xx & xx & xx & xx & xx & xx \\
%     \midrule
%     GPT-3.5-Turbo & xx & xx & xx & xx & xx & xx & xx & xx & xx & xx \\
%     +MRDA & xx & xx & xx & xx & xx & xx & xx & xx & xx & xx \\
%     +SWDA & xx & xx & xx & xx & xx & xx & xx & xx & xx & xx \\
%     \bottomrule
%     \end{tabular}
%     }
%     \caption{TODO}
%     \label{tab:results}
% \end{table*}

% \bibliography{anthology,custom,nl}
% \bibliographystyle{acl_natbib}
% \nocite{*}
\section{Bibliographical References}\label{sec:reference}

\bibliographystyle{lrec-coling2024-natbib}
\bibliography{custom,nl,anthology}

\begin{thebibliography}{22}
\expandafter\ifx\csname natexlab\endcsname\relax\def\natexlab#1{#1}\fi

\bibitem[{Al-Duleimi et~al.(2016)Al-Duleimi, Rashid, and Abdullah}]{al-duleimi-etal-2016-critical}
Hutheifa~Y. Al-Duleimi, Sabariah Hj~Md Rashid, and Ain~Nadzimah Abdullah. 2016.
\newblock A critical review of prominent theories of politeness.
\newblock \emph{Advances in Language and Literary Studies}, 7:262--270.

\bibitem[{Anderson et~al.(1991)Anderson, Bader, Bard, Boyle, Doherty, Garrod, Isard, Kowtko, McAllister, Miller et~al.}]{anderson1991hcrc}
Anne~H Anderson, Miles Bader, Ellen~Gurman Bard, Elizabeth Boyle, Gwyneth Doherty, Simon Garrod, Stephen Isard, Jacqueline Kowtko, Jan McAllister, Jim Miller, et~al. 1991.
\newblock The hcrc map task corpus.
\newblock \emph{Language and speech}, 34(4):351--366.

\bibitem[{Austin(1962)}]{austin:1962}
J.~L. Austin. 1962.
\newblock \emph{How to do things with words}.
\newblock Oxford University Press.

\bibitem[{Bender et~al.(2021)Bender, Gebru, McMillan-Major, and Shmitchell}]{bender-etal-2021-parrots}
Emily~M. Bender, Timnit Gebru, Angelina McMillan-Major, and Shmargaret Shmitchell. 2021.
\newblock On the dangers of stochastic parrots: Can language models be too big?
\newblock \emph{Proceedings of the 2021 ACM Conference on Fairness, Accountability, and Transparency}.

\bibitem[{Bender and Koller(2020)}]{bender-koller-2020-climbing}
Emily~M. Bender and Alexander Koller. 2020.
\newblock \href {https://doi.org/10.18653/v1/2020.acl-main.463} {Climbing towards {NLU}: {On} meaning, form, and understanding in the age of data}.
\newblock In \emph{Proceedings of the 58th Annual Meeting of the Association for Computational Linguistics}, pages 5185--5198, Online. Association for Computational Linguistics.

\bibitem[{Brown and Levinson(1987)}]{brown-levinson-1987-politeness}
Penelope Brown and Stephen~C Levinson. 1987.
\newblock \emph{Politeness: Some Universals in Language Usage}.
\newblock Cambridge University Press.

\bibitem[{Chung et~al.(2022)Chung, Hou, Longpre, Zoph, Tay, Fedus, Li, Wang, Dehghani, Brahma, Webson, Gu, Dai, Suzgun, Chen, Chowdhery, Castro-Ros, Pellat, Robinson, Valter, Narang, Mishra, Yu, Zhao, Huang, Dai, Yu, Petrov, Chi, Dean, Devlin, Roberts, Zhou, Le, and Wei}]{chung-etal-2022-scaling}
Hyung~Won Chung, Le~Hou, Shayne Longpre, Barret Zoph, Yi~Tay, William Fedus, Yunxuan Li, Xuezhi Wang, Mostafa Dehghani, Siddhartha Brahma, Albert Webson, Shixiang~Shane Gu, Zhuyun Dai, Mirac Suzgun, Xinyun Chen, Aakanksha Chowdhery, Alex Castro-Ros, Marie Pellat, Kevin Robinson, Dasha Valter, Sharan Narang, Gaurav Mishra, Adams Yu, Vincent Zhao, Yanping Huang, Andrew Dai, Hongkun Yu, Slav Petrov, Ed~H. Chi, Jeff Dean, Jacob Devlin, Adam Roberts, Denny Zhou, Quoc~V. Le, and Jason Wei. 2022.
\newblock \href {http://arxiv.org/abs/2210.11416} {Scaling instruction-finetuned language models}.

\bibitem[{Cohen and Levesque(1987)}]{cohen-levesque-1987-rational}
Philip~R Cohen and Hector~J Levesque. 1987.
\newblock \emph{Rational interaction as the basis for communication}, volume~87.
\newblock CSLI Stanford.

\bibitem[{Core and Allen(1997)}]{core-and-allen-1997-coding}
Mark~G. Core and James~F. Allen. 1997.
\newblock {C}oding {D}ialogs with the {DAMSL} {A}nnotation {S}cheme.

\bibitem[{Danescu-Niculescu-Mizil et~al.(2013)Danescu-Niculescu-Mizil, Sudhof, Jurafsky, Leskovec, and Potts}]{danescu-niculescu-mizil-etal-2013-computational}
Cristian Danescu-Niculescu-Mizil, Moritz Sudhof, Dan Jurafsky, Jure Leskovec, and Christopher Potts. 2013.
\newblock \href {https://aclanthology.org/P13-1025} {A computational approach to politeness with application to social factors}.
\newblock In \emph{Proceedings of the 51st Annual Meeting of the Association for Computational Linguistics (Volume 1: Long Papers)}, pages 250--259, Sofia, Bulgaria. Association for Computational Linguistics.

\bibitem[{Dutt et~al.(2020)Dutt, Joshi, and Rose}]{dutt-etal-2020-keeping}
Ritam Dutt, Rishabh Joshi, and Carolyn Rose. 2020.
\newblock \href {https://doi.org/10.18653/v1/2020.emnlp-main.605} {Keeping up appearances: Computational modeling of face acts in persuasion oriented discussions}.
\newblock In \emph{Proceedings of the 2020 Conference on Empirical Methods in Natural Language Processing (EMNLP)}, pages 7473--7485, Online. Association for Computational Linguistics.

\bibitem[{Grice(1975)}]{grice:1975}
Herbert~Paul Grice. 1975.
\newblock Logic and conversation.
\newblock In P.~Cole and J.~Morgan, editors, \emph{Syntax and semantics, vol 3}. Academic Press, New York.

\bibitem[{Hancher(1979)}]{hancher-1979-classification}
Michael Hancher. 1979.
\newblock The classification of cooperative illocutionary acts.
\newblock \emph{Language in society}, 8(1):1--14.

\bibitem[{He et~al.(2021)He, Tavabi, Lerman, and Soleymani}]{he-etal-2021-speaker-turn}
Zihao He, Leili Tavabi, Kristina Lerman, and Mohammad Soleymani. 2021.
\newblock \href {https://doi.org/10.18653/v1/2021.findings-emnlp.185} {Speaker turn modeling for dialogue act classification}.
\newblock In \emph{Findings of the Association for Computational Linguistics: EMNLP 2021}, pages 2150--2157, Punta Cana, Dominican Republic. Association for Computational Linguistics.

\bibitem[{Murzaku et~al.(2022)Murzaku, Zeng, Markowska, and Rambow}]{murzaku-etal-2022-examining}
John Murzaku, Peter Zeng, Magdalena Markowska, and Owen Rambow. 2022.
\newblock \href {https://aclanthology.org/2022.coling-1.66} {Re-examining {F}act{B}ank: Predicting the author{'}s presentation of factuality}.
\newblock In \emph{Proceedings of the 29th International Conference on Computational Linguistics}, pages 786--796, Gyeongju, Republic of Korea. International Committee on Computational Linguistics.

\bibitem[{Purkarthofer and Flubacher(2022)}]{purkarthofer-flubacher-2022-speaking}
Judith Purkarthofer and Mi-Cha Flubacher. 2022.
\newblock \emph{Speaking Subjects in Multilingualism Research: Biographical and Speaker-centred Approaches}, volume~7.
\newblock Channel View Publications.

\bibitem[{Shriberg et~al.(2004)Shriberg, Dhillon, Bhagat, Ang, and Carvey}]{shriberg-etal-2004-icsi}
Elizabeth Shriberg, Raj Dhillon, Sonali Bhagat, Jeremy Ang, and Hannah Carvey. 2004.
\newblock \href {https://aclanthology.org/W04-2319} {The {ICSI} meeting recorder dialog act ({MRDA}) corpus}.
\newblock In \emph{Proceedings of the 5th {SIG}dial Workshop on Discourse and Dialogue at {HLT}-{NAACL} 2004}, pages 97--100, Cambridge, Massachusetts, USA. Association for Computational Linguistics.

\bibitem[{Stolcke et~al.(2000{\natexlab{a}})Stolcke, Ries, Coccaro, Shriberg, Bates, Jurafsky, Taylor, Martin, Ess-Dykema, and Meteer}]{S*:2000}
Andreas Stolcke, Klaus Ries, Noah Coccaro, Elizabeth Shriberg, Rebecca Bates, Daniel Jurafsky, Paul Taylor, Rachel Martin, Carol~Van Ess-Dykema, and Marie Meteer. 2000{\natexlab{a}}.
\newblock Dialogue act modeling for automatic tagging and recognition of conversational speech.
\newblock \emph{Computational Linguistics}, 26:339--373.

\bibitem[{Stolcke et~al.(2000{\natexlab{b}})Stolcke, Ries, Coccaro, Shriberg, Bates, Jurafsky, Taylor, Martin, Van Ess-Dykema, and Meteer}]{stolcke-etal-2000-dialogue}
Andreas Stolcke, Klaus Ries, Noah Coccaro, Elizabeth Shriberg, Rebecca Bates, Daniel Jurafsky, Paul Taylor, Rachel Martin, Carol Van Ess-Dykema, and Marie Meteer. 2000{\natexlab{b}}.
\newblock \href {https://aclanthology.org/J00-3003} {Dialogue act modeling for automatic tagging and recognition of conversational speech}.
\newblock \emph{Computational Linguistics}, 26(3):339--374.

\bibitem[{Walker et~al.(1997)Walker, Cahn, and Whittaker}]{walker-etal-1997-improvising}
Marilyn~A. Walker, Janet~E. Cahn, and Stephen~J. Whittaker. 1997.
\newblock \href {https://doi.org/10.1145/267658.267680} {Improvising linguistic style: Social and affective bases for agent personality}.
\newblock In \emph{Proceedings of the First International Conference on Autonomous Agents}, AGENTS '97, page 96–105, New York, NY, USA. Association for Computing Machinery.

\bibitem[{Wang et~al.(2019)Wang, Shi, Kim, Oh, Yang, Zhang, and Yu}]{wang-etal-2019-persuasion}
Xuewei Wang, Weiyan Shi, Richard Kim, Yoojung Oh, Sijia Yang, Jingwen Zhang, and Zhou Yu. 2019.
\newblock \href {https://doi.org/10.18653/v1/P19-1566} {Persuasion for good: Towards a personalized persuasive dialogue system for social good}.
\newblock In \emph{Proceedings of the 57th Annual Meeting of the Association for Computational Linguistics}, pages 5635--5649, Florence, Italy. Association for Computational Linguistics.

\bibitem[{Zhang et~al.(2021)Zhang, Li, Deng, Bing, and Lam}]{zhang-etal-2021-towards-generative}
Wenxuan Zhang, Xin Li, Yang Deng, Lidong Bing, and Wai Lam. 2021.
\newblock \href {https://doi.org/10.18653/v1/2021.acl-short.64} {Towards generative aspect-based sentiment analysis}.
\newblock In \emph{Proceedings of the 59th Annual Meeting of the Association for Computational Linguistics and the 11th International Joint Conference on Natural Language Processing (Volume 2: Short Papers)}, pages 504--510, Online. Association for Computational Linguistics.

\end{thebibliography}

% \section{Language Resource References}
% \label{lr:ref}
% \bibliographystylelanguageresource{lrec-coling2024-natbib}
% \bibliographylanguageresource{languageresource}

\end{document}